\documentclass{bmvc2k}

\newcommand{\papertitle}{SWARD: Stochastic Window-Attention-\\Based Relational Distillation for Cross-Architectural Semantic Segmentation}
\newcommand{\papertitleshort}{Stochastic Window Attention Relational Distillation}
\title{\papertitle}

\addauthor{Aditya Makineni}{amakinen@uab.edu}{1}
\addauthor{Qing Tian$^{*}$}{qtian@uab.edu}{1}

\addinstitution{
 Department of Computer Science\\
 University of Alabama at Birmingham\\
 Birmingham, US\\[2pt]
 $^{*}$Corresponding author.
}

\runninghead{Makineni \& Tian}{\papertitleshort}

\usepackage{amssymb}
\usepackage{multirow}
\usepackage{comment}
\usepackage{colortbl}
\usepackage{xcolor}
\usepackage{float}
\usepackage{subfigure}
\usepackage{fontawesome}
\usepackage{tikz}

\newcommand{\faPentagon}{\tikz[baseline=-0.4ex]\fill
  (90:0.45ex) -- (162:0.45ex) -- (234:0.45ex) -- (306:0.45ex) -- (18:0.45ex) -- cycle;}

\definecolor{roadc}{HTML}{1F77B4}
\definecolor{sidewalkc}{HTML}{E377C2}
\definecolor{vegetationc}{HTML}{795C34}
\definecolor{terrainc}{HTML}{BCBD22}
\definecolor{personc}{HTML}{E6194B}
\definecolor{riderc}{HTML}{FFE119}
\definecolor{truckc}{HTML}{F58231}
\definecolor{busc}{HTML}{911EB4}
\definecolor{motorcyclec}{HTML}{F032E6}
\definecolor{bicyclec}{HTML}{7FFFFD}

\newcommand{\mkroad}{\textcolor{roadc}{\tiny\faCircle}}
\newcommand{\mksidewalk}{\textcolor{sidewalkc}{\tiny\faCircle}}
\newcommand{\mkterrain}{\textcolor{terrainc}{\tiny\faCircle}}
\newcommand{\mkvegetation}{\textcolor{vegetationc}{\Large\faPentagon}}
\newcommand{\mkperson}{\textcolor{personc}{\footnotesize\faTimes}}
\newcommand{\mkrider}{\textcolor{riderc}{\footnotesize\faTimes}}
\newcommand{\mktruck}{\textcolor{truckc}{\footnotesize\faTimes}}
\newcommand{\mkbus}{\textcolor{busc}{\footnotesize\faTimes}}
\newcommand{\mkmotorcycle}{\textcolor{motorcyclec}{\footnotesize\faTimes}}
\newcommand{\mkbicycle}{\textcolor{bicyclec}{\footnotesize\faTimes}}
\begin{document}

\maketitle

\begin{abstract}
Large-scale vision foundation models have driven substantial gains on dense prediction tasks such as semantic segmentation, but their size makes deployment impractical in resource-constrained settings, motivating knowledge distillation as a means of transferring their capabilities to lightweight student networks. However, modern foundation teachers are predominantly transformer-based models that encode global context, whereas efficient students are typically convolutional networks with locally biased receptive fields. Existing distillation methods largely assume architectural homogeneity and rely on direct feature mimicry, which fails to bridge this representational gap and neglects the structured spatial dependencies and discriminative organization required for accurate semantic segmentation. In this paper, we propose SWARD, a knowledge distillation framework that addresses this gap through two complementary mechanisms. First, we introduce a Multi-Scale Windowed Attention Distillation (MWAD) module that aligns teacher-student attention-based relations within stochastically shifted window partitions whose offsets are randomly resampled at every training iteration. This removes window boundary bias, and, combined with the multi-scale design, captures both short- and long-range spatial dependencies. Second, we introduce Prototype Discriminative Regularization (PDR), a loss that helps shape the student's feature distribution by enforcing inter-class separation and intra-class compactness, further sharpening the discriminative structure beyond what feature mimicry alone can produce under the student's reduced capacity. Experiments across different vision applications (i.e., urban scene parsing and medical image segmentation) show that SWARD achieves state-of-the-art performance.
\end{abstract}

\section{Introduction}


\begin{figure}[t]
    \centering
    \begin{minipage}[t]{0.26\textwidth}
        \centering
        \includegraphics[height=2cm]{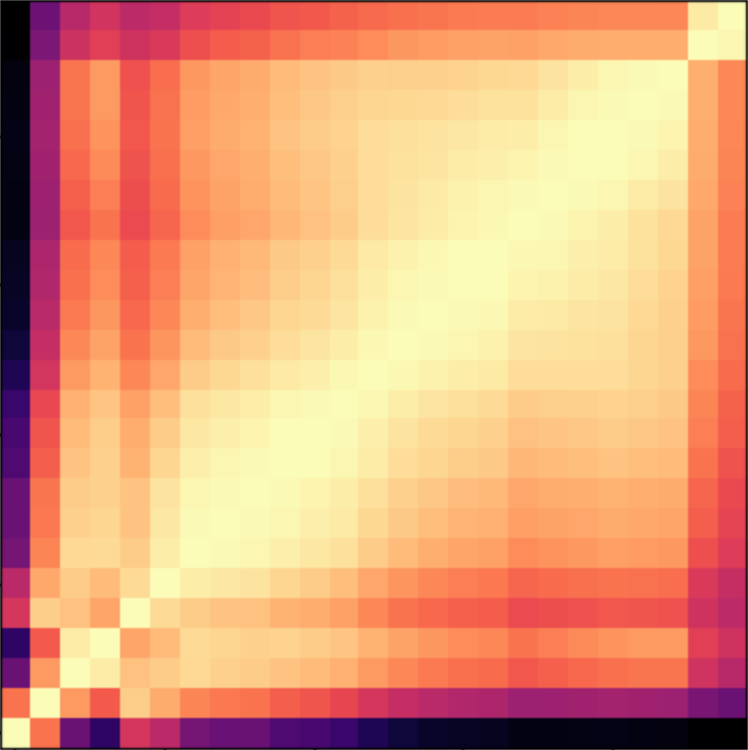}\\[-2.5pt]
        {\scriptsize (a) Transformer / Transformer}
    \end{minipage}
    \begin{minipage}[t]{0.26\textwidth}
        \centering
        \includegraphics[height=2cm]{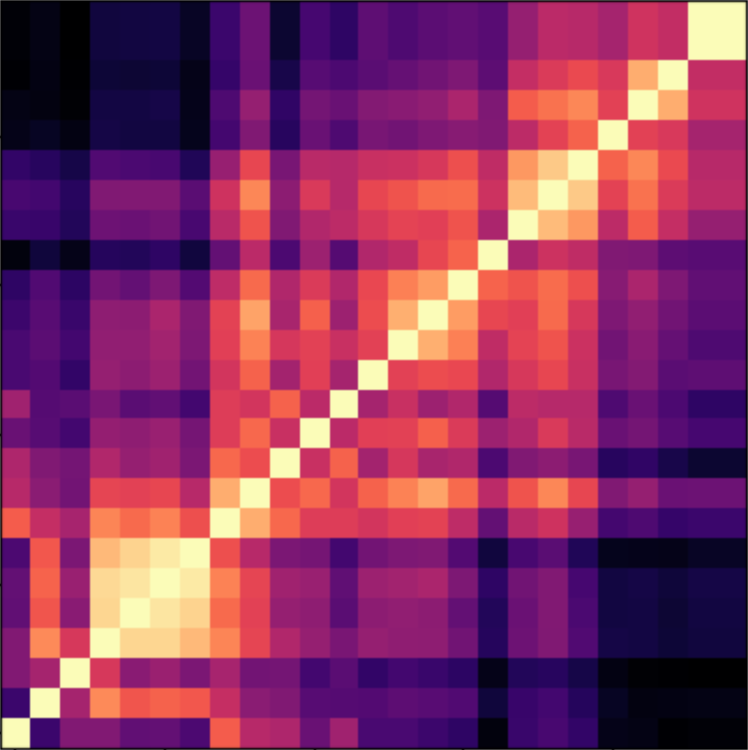}\\[-2.5pt]
        {\scriptsize (b) CNN / CNN}
    \end{minipage}
    \begin{minipage}[t]{0.26\textwidth}
        \centering
        \includegraphics[height=2cm]{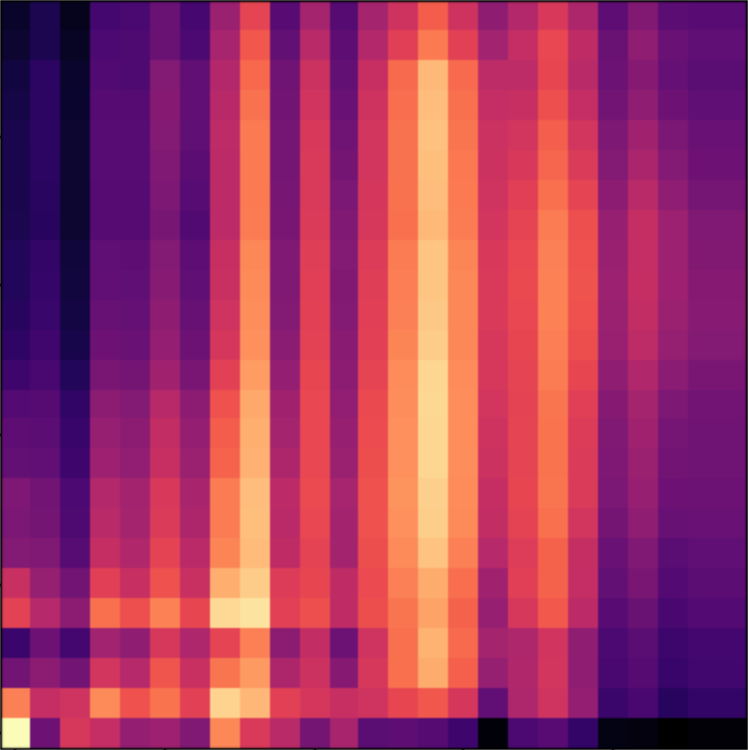}\\[-2.5pt]
        {\scriptsize (c) Transformer / CNN}
    \end{minipage}
    \hspace{0.4em}
    \begin{minipage}[t]{0.08\textwidth}
        \centering
        \includegraphics[height=2cm, trim=0 8.5 0 7, clip]{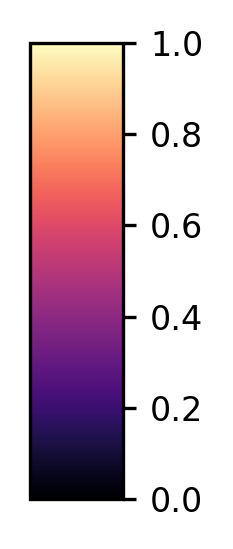}\\[-2.5pt]
    \end{minipage}
    \vspace{1.5pt}
    \caption{Representational gap between heterogeneous architectures. Layer-wise Centered Kernel Alignment (CKA) between all layer pairs of the two networks named on each panel's axes: (a) teacher against itself, (b) student against itself, (c) teacher against student. The self-comparisons show clear diagonal correspondence, whereas the cross-architecture map does not, indicating that teacher and student layers admit no consistent alignment. Teacher: SAM2.1 Hiera-L; student: MobileNetV3-Small.}
    \label{fig:cka_comparison}
\end{figure}

\label{sec:intro}
Large-scale vision foundation models achieve strong dense-prediction performance through massive scale, spanning tasks like semantic segmentation and object detection \cite{ryali2023hiera, ravi2025sam, zhai2022scaling}, but their computational footprint limits deployment in resource-constrained applications such as autonomous driving and medical imaging, motivating knowledge distillation (KD) \cite{hinton2015distilling, huang2023knowledge, 2014arXiv1412.6550R} from a heavy teacher to a lightweight student. Dense-prediction KD is more challenging than its classification counterpart: pixels carry uneven semantic weight \cite{liu2024bpkd}, and naively aggregated pixel-wise mimicry losses can dilute the structured dependencies needed for accurate local predictions \cite{liu2019structured, e25010125}.
Subsequent works improve over uniform aggregation through region-aware strategies \cite{yang2022focal, zheng2022localization}, yet a more fundamental limitation persists: existing dense-prediction KD methods largely assume architectural homogeneity between teacher and student, where intermediate features are spatially and semantically aligned by design. In practice, however, the most powerful teachers increasingly rely on transformer-based or hybrid foundation architectures that encode long-range dependencies and token-wise relational structure through self-attention \cite{dosovitskiy2020image, liu2021swin}, while efficient students are typically convolutional networks optimized for low-latency deployment \cite{sandler2018mobilenetv2, howard2019searching, tan2019efficientnet}, whose hierarchical receptive fields capture only locally biased features \cite{raghu2021vision, cordonnier2019relationship}. This architectural mismatch makes knowledge transfer more difficult, limiting the effectiveness of direct feature mimicry.
To quantify this architectural gap, we perform a layer-wise Centered Kernel Alignment (CKA) \cite{kornblith2019similarity} analysis comparing intermediate representations of a SAM2.1 Transformer teacher \cite{ravi2025sam} and a MobileNet CNN student \cite{howard2019searching}. As shown in Figure \ref{fig:cka_comparison}, the results reveal a substantial representational gap between Transformer and CNN features.

\begin{figure*}[t]
    \centering

    \makebox[0.23\textwidth][c]{\small Input Image}\hfill
    \makebox[0.23\textwidth][c]{\small Teacher}\hfill
    \makebox[0.23\textwidth][c]{\small Baseline Student}\hfill
    \makebox[0.23\textwidth][c]{\small SWARD (Ours)}
    
    \par\vspace{-2pt}

    \makebox[0.23\textwidth][c]{\subfigure{\includegraphics[width=0.21\textwidth, trim={150pt 100pt 500pt 0}, clip]{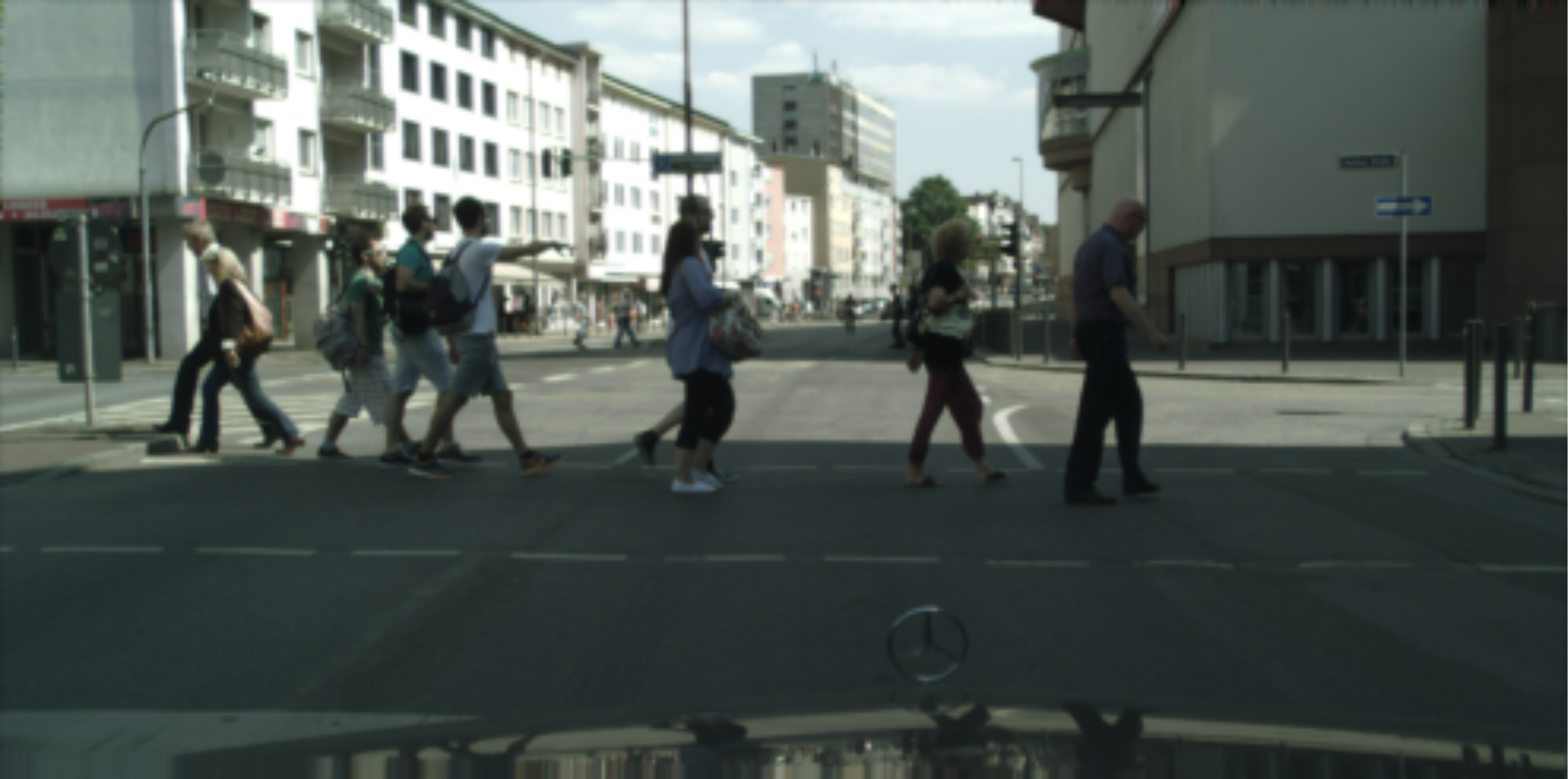}}}\hfill
    \makebox[0.23\textwidth][c]{\subfigure{\includegraphics[width=0.21\textwidth, trim={150pt 100pt 500pt 0}, clip]{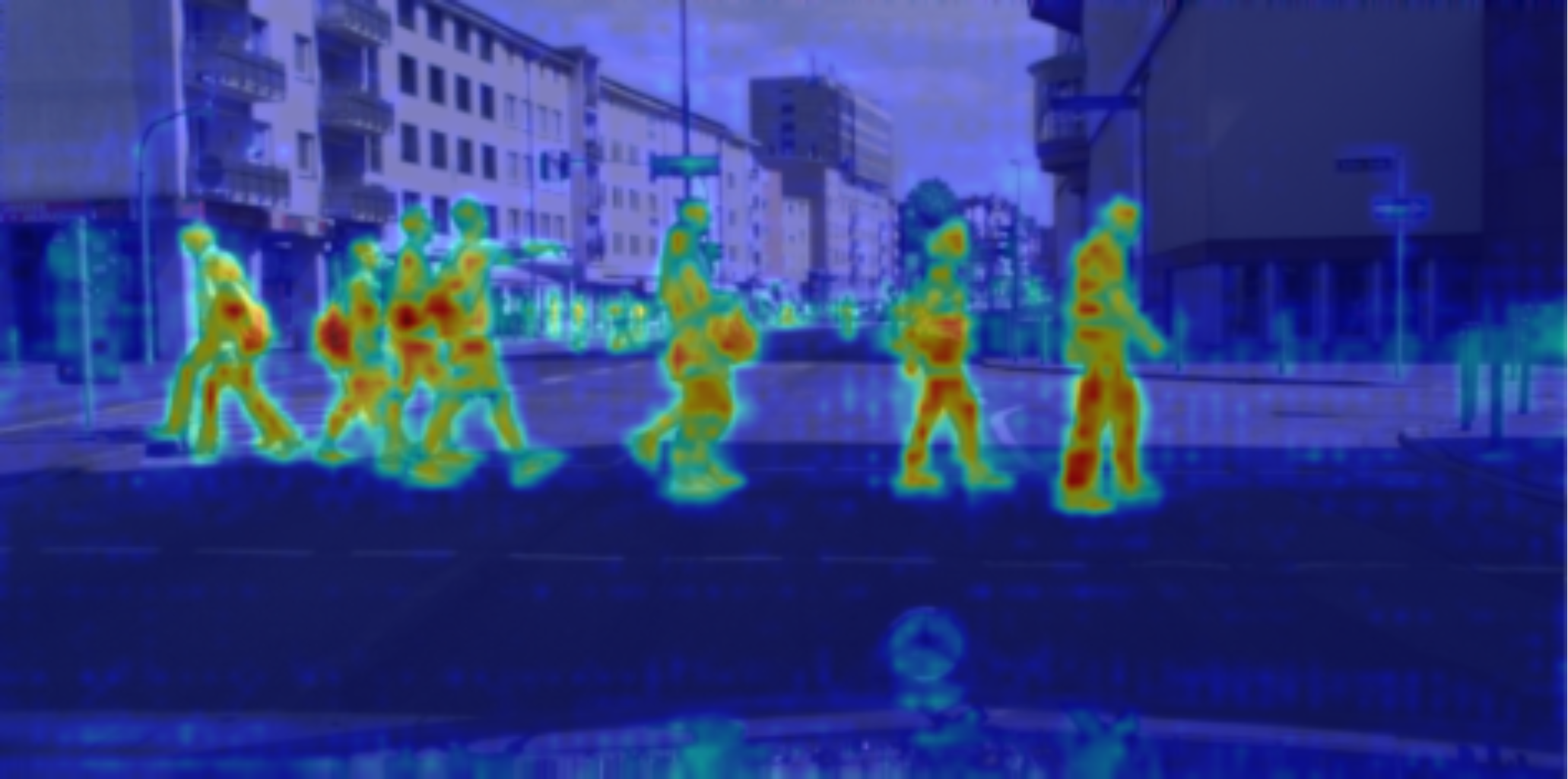}}}\hfill
    \makebox[0.23\textwidth][c]{\subfigure{\includegraphics[width=0.21\textwidth, trim={150pt 100pt 500pt 0}, clip]{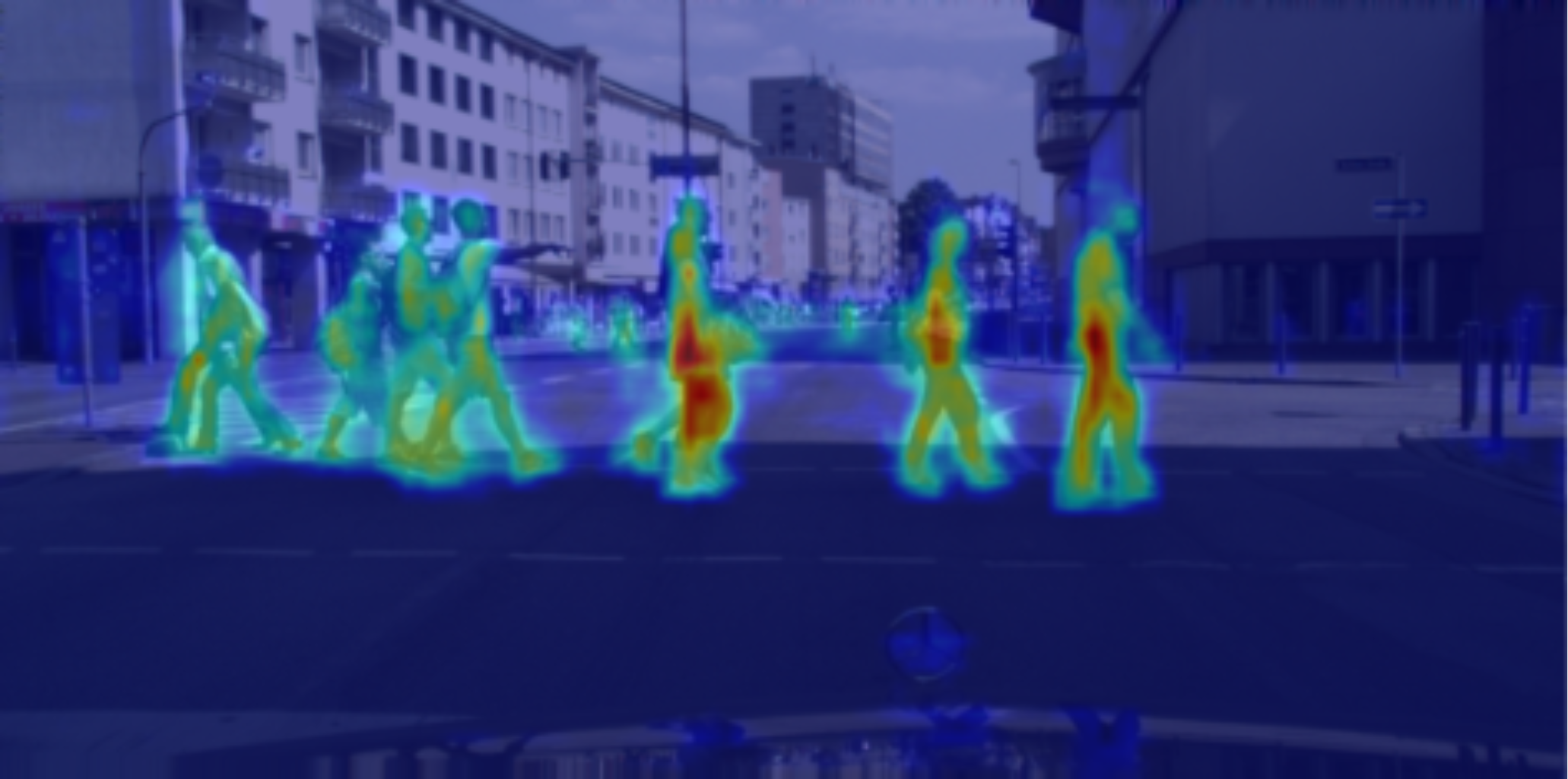}}}\hfill
    \makebox[0.23\textwidth][c]{\subfigure{\includegraphics[width=0.21\textwidth, trim={150pt 100pt 500pt 0}, clip]{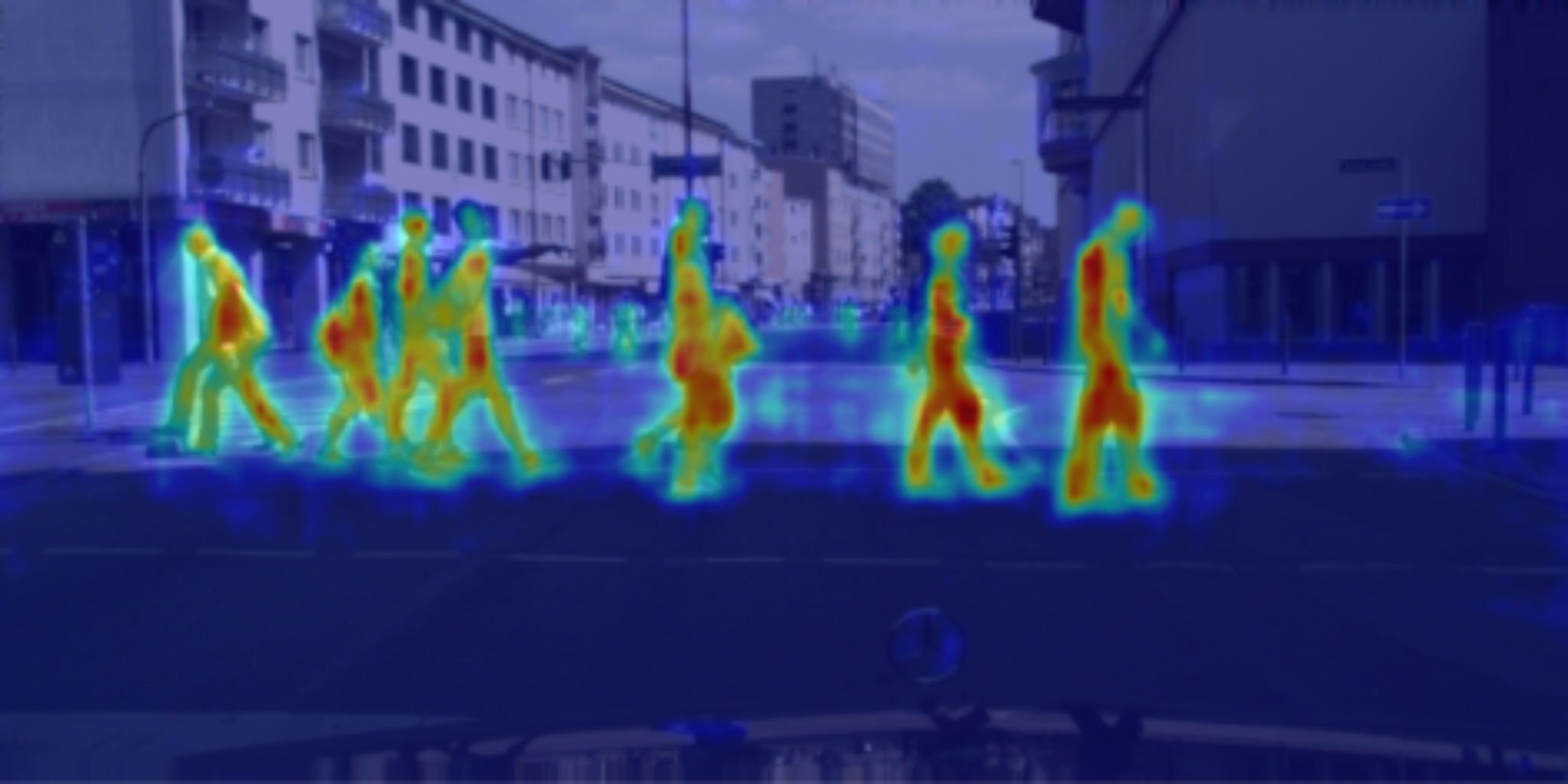}}}
    
    \par\vspace{-10pt} 

    \makebox[0.23\textwidth][c]{\subfigure{\includegraphics[width=0.16\textwidth, trim={10pt 25pt 25pt 60pt}, clip]{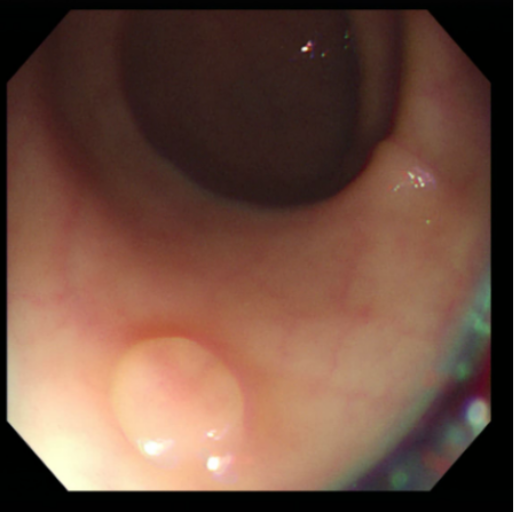}}}\hfill
    \makebox[0.23\textwidth][c]{\subfigure{\includegraphics[width=0.16\textwidth, trim={10pt 25pt 25pt 60pt}, clip]{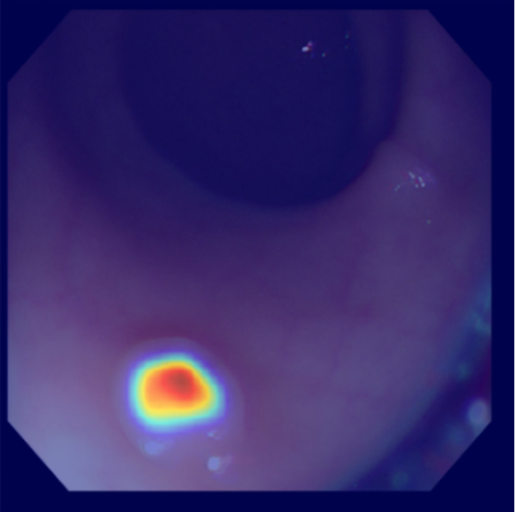}}}\hfill
    \makebox[0.23\textwidth][c]{\subfigure{\includegraphics[width=0.16\textwidth, trim={10pt 25pt 25pt 60pt}, clip]{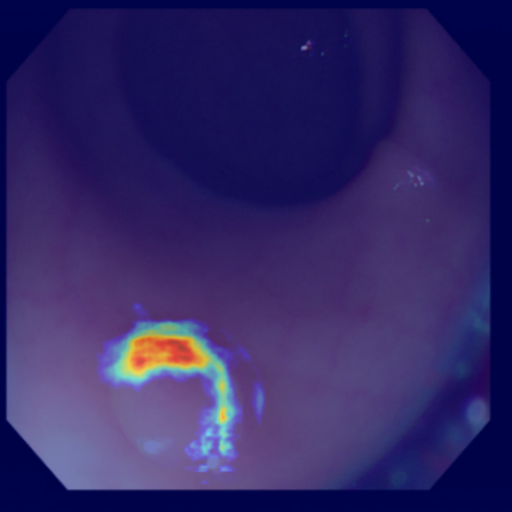}}}\hfill
    \makebox[0.23\textwidth][c]{\subfigure{\includegraphics[width=0.16\textwidth, trim={10pt 25pt 25pt 60pt}, clip]{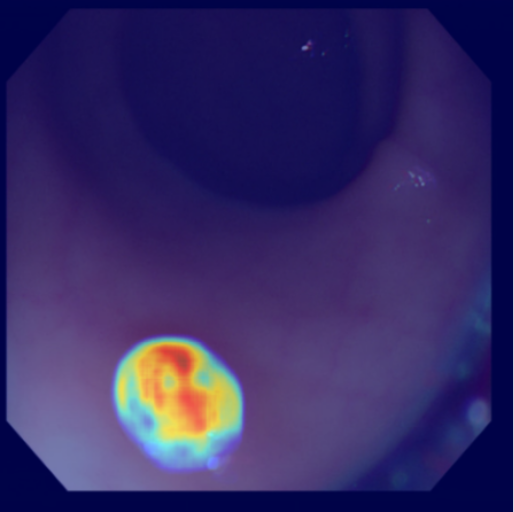}}}
    \vspace{-6pt}
    \caption{Grad-CAM maps across different vision applications: urban-scene (Cityscapes, top) and medical (CVC-ColonDB, bottom). Columns from left to right: input image, teacher, baseline student, and +SWARD (ours). SWARD yields more complete and better-localized coverage of the target objects, closely matching the teacher. Images are cropped for visibility.}
    \label{fig:gradcam_comparison}
\end{figure*}

These challenges are most pronounced in semantic segmentation, where exhaustive per-pixel classification demands both spatially structured knowledge transfer and a highly discriminative feature space. To address them, we propose Stochastic Window-Attention-Based Relational Distillation (SWARD), a knowledge distillation framework for heterogeneous teacher-student architectures. Rather than directly enforcing feature-level mimicry between mismatched architectures, SWARD introduces a mechanism to reconcile the global contextual nature of transformer-based foundation models with the local hierarchical bias of CNNs. Specifically, we introduce a Multi-Scale Windowed Attention Distillation (MWAD) module that decomposes feature transfer into attention, semantic structure, and contextual alignment, enabling the student to inherit both the spatial reasoning and semantic structure of the teacher without requiring architectural homogeneity. As shown in Figure \ref{fig:gradcam_comparison}, this attention-centric transfer allows the student to focus better on target objects. While the baseline student often attends to the target weakly or only in part, our distilled student achieves much more complete coverage, closely matching the teacher's. MWAD aligns these spatial attention maps, but the student's smaller capacity limits how successfully relational mimicry can approximate the teacher's class-wise separability. To address this, we further introduce Prototype Discriminative Regularization (PDR), which makes the student's class clusters more compact and better separated than relation matching alone achieves, sharpening the decision boundaries on which accurate segmentation depends. Together, these components enable SWARD to effectively transfer knowledge from attention-based teachers to efficient convolutional students.

\section{Related Work}
\label{sec:related_works}
\subsection{Heterogeneous Knowledge Distillation}
Knowledge distillation for dense prediction is typically developed under the assumption that teacher and student networks share compatible architectures, allowing intermediate representations to be aligned through direct feature matching. Beyond pixel-wise feature reconstruction \cite{2014arXiv1412.6550R}, later methods transfer higher-level structure of the teacher's representation rather than its raw activations: attention transfer \cite{zagoruyko2017paying} matches activation-based spatial attention maps that mark where a CNN concentrates its responses within an image, while cross-image relational distillation (CIRKD) \cite{yang2022cross} transfers pixel relations across images for segmentation. Other dense-prediction methods concentrate transfer on informative regions \cite{yang2022focal, zheng2022localization} or impose structured spatial constraints \cite{liu2019structured}. Although relational objectives transfer more robustly than absolute feature values, these methods assume architecturally comparable teacher and student features and tend to perform best when the two feature spaces correspond.

Recent work on heterogeneous knowledge distillation relaxes this assumption by considering teacher and student models with fundamentally different architectures. Cross-Archi\-tec\-ture Knowledge Distillation (CKD) \cite{liu2022cross} addresses this challenge by projecting student features into the teacher's attention and feature spaces, where the mismatched representations become comparable, and training the student to reproduce the teacher's projected representations. However, this alignment operates at a single, coarse token resolution with globally computed attention: developed for classification, it neither localizes the relational targets nor spans multiple scales, so the fine structure of object boundaries and small regions is only weakly constrained. This can be limiting for semantic segmentation, where a label is assigned at every pixel and accuracy depends on exactly this spatial detail. Heterogeneous Knowledge Distillation using Information Flow Modeling \cite{passalis2020heterogeneous} transfers how information is progressively transformed across layers rather than raw activations but represents each layer by pairwise similarities between transfer samples, leaving multi-scale spatial structure unconstrained. OFA-KD \cite{hao2023one} bridges heterogeneous architectures by projecting intermediate features into a shared logit space; however, aligning in this output space discards the high-resolution spatial details that pixel-level prediction depends on. Most closely related, HMKD \cite{xu2025hmkd} aligns patch-level self-attention between Transformer and CNN segmentation models, but its relational transfer matches only the attention map within a fixed patch partition, and its authors report that mixing patch sizes degrades accuracy. 

For heterogeneous distillation in semantic segmentation, these limitations motivate relational alignment that is simultaneously localized, spans multiple scales, and is not tied to a single fixed partition. While deterministic shifted-window attention has proven effective for efficient representation learning \cite{liu2021swin}, using shifted-window attention as a relational distillation target across architectures, particularly with stochastic shifts, remains unexplored, a gap our MWAD module addresses by transferring attention, value, and context relations across multiple window scales under stochastically resampled shifts.

\subsection{Discriminative Feature Regularization for Dense Prediction}
Beyond aligning student features to the teacher, the geometric organization of the student's own feature space strongly influences how well classes can be separated, a property that standard distillation objectives rarely target directly. Feature regression or similarity matching encourages the student to imitate teacher representations but does not explicitly shape inter-class separation or intra-class compactness, often yielding overlapping class embeddings and blurred decision boundaries. Three lines of work add such structure explicitly. Contrastive methods model relationships between pixel or region features, pulling same-class embeddings together while pushing different-class ones apart \cite{wang2021exploring}, and have been adapted to distillation for dense prediction \cite{fan2023augmentation}; however, they shape the embedding through pairwise sample contrast rather than directly constraining the student's class-wise geometry. Prototype-based methods instead summarize each class by a feature centroid and pull its members toward that centroid, producing more compact and semantically consistent clusters \cite{CHEN2024117186}, but they have been studied mainly in homogeneous or self-distillation settings rather than the cross-architecture case, where the student must organize its space while aligning to a structurally different teacher. Discriminant-maximization methods directly maximize between-class separation while minimizing within-class scatter, improving accuracy and robustness and guiding compression in dense prediction such as object detection \cite{choi2025improving, lan2026visual}, but they have largely been applied outside knowledge distillation.

Across these contrastive, prototype, and discriminant approaches, discriminative regularization has rarely been coupled with knowledge distillation, and where it has, only between architecturally compatible teacher-student pairs, leaving the cross-architecture setting underexplored. Our Prototype Discriminative Regularization (PDR) targets this gap by folding discriminative feature regularization directly into heterogeneous distillation, so that the student both matches the teacher's relations and maintains a separable, well-organized embedding space throughout transfer.

\section{Methodology}
\label{sec:proposed_method}

\begin{figure}[t]
\centering
\includegraphics[width=0.8\textwidth]{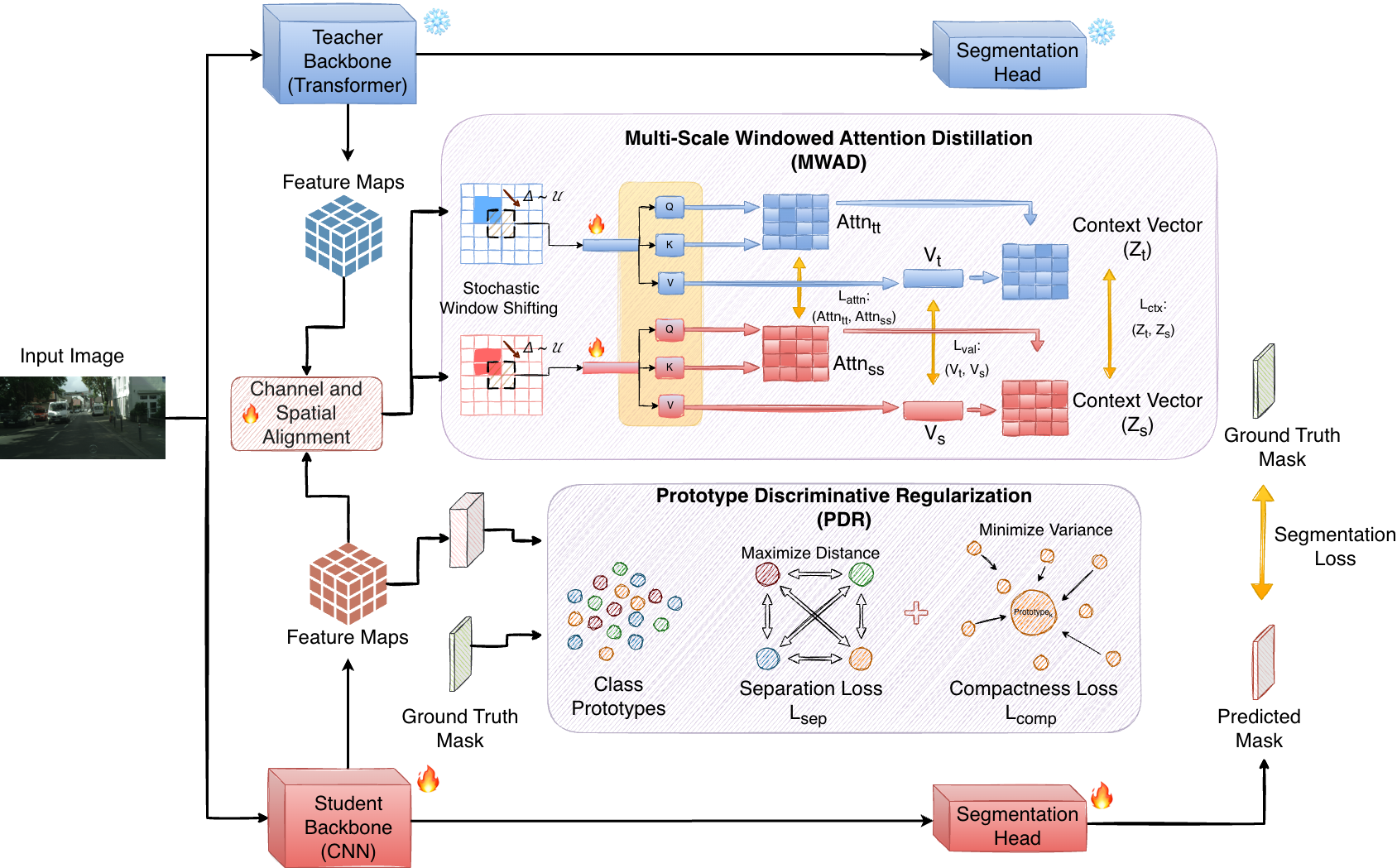}
\par\vspace{-6pt}
\caption{Overview of SWARD. A frozen attention-based teacher (top, blue) and a trainable convolutional student (bottom, red) extract multi-scale features. MWAD aligns teacher and student features at corresponding stages, partitions them into stochastically shifted local windows, and transfers the teacher's relational structure through attention, value, and context losses. PDR forms per-class prototypes from student features and enforces inter-class separability and intra-class compactness via separation and compactness losses. The snowflake/flame denotes frozen/trainable components.}
\label{fig:kd_overview}
\end{figure}
We propose SWARD (Figure \ref{fig:kd_overview}), a heterogeneous knowledge distillation framework that transfers structured knowledge from a high-capacity attention-based teacher to an efficient convolutional student. We formulate distillation as a structured relational transfer problem that aligns how features interact spatially and semantically, rather than forcing direct activation-level mimicking.

Given an input image $x$, the student network $\mathcal{S}$ produces a dense prediction map $\hat{y}$ and extracts multi-scale intermediate feature maps $\{f^i_s\}^N_{i=1}$ at $N$ stages, while a pretrained teacher network $\mathcal{T}$ extracts feature maps $\{f^i_t\}^N_{i=1}$ at the corresponding scales. The teacher and student are architecturally heterogeneous: the attention-based teacher encodes global, relational structure, whereas the convolutional student captures locally biased features, often differing in channel and spatial dimensionality. Our objective is to transfer the teacher's structural-spatial and semantic knowledge into the student while preserving the efficiency of the student architecture at inference time. To address this, SWARD consists of two components: Multi-Scale Windowed Attention Distillation (MWAD), which performs localized relational alignment across multiple resolutions, and Prototype Discriminative Regularization (PDR), which regularizes the student feature space to improve intra-class compactness and inter-class separability.

\subsection{Multi-Scale Windowed Attention Distillation (MWAD)}
The goal of this module is to transfer structured knowledge between heterogeneous feature spaces. Instead of aligning raw feature activations, we align the relational structure produced by windowed attention interactions across multiple spatial scales.

\subsubsection{Multi-Scale Feature Alignment}
Let $f^i_t \in \mathbb{R}^{C^i_t \times H^i_t \times W^i_t}$ and $f^i_s \in \mathbb{R}^{C^i_s \times H^i_s \times W^i_s}$ denote the teacher and student feature maps at stage $i$. When the two networks do not expose the same number of stages, we still form $N$ teacher-student pairs at comparable depths, selecting the corresponding subset from the network with more stages. Since the paired feature maps may still differ in both spatial and channel dimensionality, we align them before distillation.
To resolve channel mismatch, the teacher feature is projected into the student channel space through a learnable linear projection, $\tilde{f}^i_t = \phi(f^i_t)$, where $\phi(\cdot)$ is a $1 \times 1$ convolution that maps the teacher's $C^i_t$ channels to the student's $C^i_s$ channels. To resolve the spatial mismatch, both teacher and student feature maps are resized to a common resolution $(\hat{f}^i_t, \hat{f}^i_s) = \psi(\tilde{f}^i_t, f^i_s)$, where $\psi(\cdot)$ denotes bilinear resampling of both feature maps to the minimum shared spatial resolution:
$\hat{H}_i, \hat{W}_i = \min(H^i_t, H^i_s), \min(W^i_t, W^i_s)$. This yields aligned teacher and student features $\hat{f}^i_t, \hat{f}^i_s \in \mathbb{R}^{C^i_s \times \hat{H}_i \times \hat{W}_i}$.

\subsubsection{Stochastically Shifted Window Partitioning}
\label{subsec:window_par}
Dense-prediction KD requires preserving relational and structural consistency across multiple spatial granularities. Rather than relating all spatial positions globally, which scales quadratically with the feature resolution and reaches far beyond the local receptive fields of the convolutional student, we adopt a multi-scale windowed partitioning strategy, where each aligned feature map is decomposed into non-overlapping windows of different sizes to capture spatial relations at varying receptive fields. For a stage $i$ with window size $M_i \times M_i$, each aligned feature map is divided into
$\mathcal{W}(\hat{f}^i) = \{w^i_n\}^{N_w}_{n=1}$, where each window $w^i_n \in \mathbb{R}^{C^i_s \times M_i \times M_i}$ and $N_w = \frac{\hat{H}_i\hat{W}_i}{M^2_i}$. When the feature map dimensions are not perfectly divisible by the window size $M_i$, standard zero-padding is applied to the right and bottom boundaries.

Convolution naturally captures local patterns but not in the same relational form as attention. Within each window, partitioning gives the teacher and student a common set of spatial tokens, allowing the teacher's token-to-token attention and the student's convolutional aggregation to be compared in the same spatial neighborhood.
While fixed windows efficiently localize computation, they introduce artificial boundaries that limit cross-regional interaction. To mitigate this, we introduce stochastically shifted window partitioning. Unlike the deterministic offset used in the Swin Transformer's shifted-window attention \cite{liu2021swin}, we draw an independent offset for each scale $i$ at every training iteration:
\begin{equation}(\Delta x_i, \Delta y_i) \sim \mathcal{U}\{-M_i, \ldots, M_i\}.
\end{equation}We then apply cyclic shifts to the corresponding feature map:
\begin{equation}\hat{f}^{i\prime} = \mathrm{Roll}(\hat{f}^i, -\Delta x_i, -\Delta y_i).
\end{equation}This produces stochastically varying window configurations across both training iterations and scales, allowing the model to capture cross-boundary dependencies. All subsequent per-window computations operate on the shifted maps; we omit the prime and window index for notational simplicity.

\subsubsection{Relational QKV-Based Distillation}
Since the convolutional student does not natively expose attention queries, keys, and values, we project each network's output features into a shared relational basis using lightweight learned projections:
\begin{equation}Q^i_t, K^i_t, V^i_t = \Theta^i_t(\hat{f}^i_t), \quad Q^i_s, K^i_s, V^i_s = \Theta^i_s(\hat{f}^i_s),
\end{equation}
\noindent where $\Theta^i_t$ and $\Theta^i_s$ denote independent $1 \times 1$ convolutional projections.
These projected embeddings are used to compute local scaled attention:
\begin{equation}A^i_t = \mathrm{Softmax}(\frac{Q^i_t(K^i_t)^\intercal}{\sqrt{d_i}}), \quad A^i_s = \mathrm{Softmax}(\frac{Q^i_s(K^i_s)^\intercal}{\sqrt{d_i}}).
\end{equation}The attention matrices encode pairwise relationships between spatial tokens within each local window, capturing how strongly each token attends to the others.
Contextual representations are then obtained as $Z^i_t = A^i_tV^i_t, \hspace{0.5em} Z^i_s = A^i_sV^i_s$. These context vectors aggregate information from all tokens in the window according to the learned attention weights, encoding both appearance and spatial dependencies.

At each stage, we apply three complementary matching losses on attention, value, and context, targeting the relational pattern, per-token semantics, and aggregated output of the teacher's attention computation.
Specifically, the attention loss matches the teacher's pairwise attention pattern within each window via KL divergence over the attention distributions, transferring the teacher's relational structure:
\begin{equation}\mathcal{L}^i_{attn} = \mathrm{KL}(A^i_t \| A^i_s).
\end{equation}
The value loss uses cosine distance to align the direction of per-token value embeddings, transferring the teacher's pre-mixing feature content:
\begin{equation}\mathcal{L}^i_{val} = \frac{1}{M_i^2}\sum_{j=1}^{M_i^2}\left(1 - \frac{\langle V^i_{s,j}, V^i_{t,j} \rangle}{\|V^i_{s,j}\|_2\,\|V^i_{t,j}\|_2}\right),
\end{equation}
\noindent where $V^i_{\cdot,j}$ denotes the value embedding of token $j$ within the window.
The context loss penalizes the squared Frobenius distance between teacher and student context vectors, transferring how the teacher aggregates information within each window:
\begin{equation}\mathcal{L}^i_{ctx} = \|Z^i_s - Z^i_t\|_F^2.
\end{equation}
The three terms are combined into the per-stage objective:
\begin{equation}\mathcal{L}^i_{MWAD} = \mathcal{L}^i_{attn} + \mathcal{L}^i_{val} + \mathcal{L}^i_{ctx}.
\end{equation}The multi-scale distillation loss aggregates these objectives across all $N$ stages:
\begin{equation}\mathcal{L}_{MWAD} = \frac{1}{N}\sum_{i=1}^{N} \mathcal{L}^i_{MWAD}.
\end{equation}
\subsection{Prototype Discriminative Regularization (PDR)}

Accurate semantic segmentation depends on a discriminative feature space: strong inter-class separation sharpens decision boundaries between object categories, while intra-class compactness yields consistent labeling within each region. Cross-architectural distillation faces a capacity gap that relation transfer alone cannot close: foundation-scale teachers carve cleanly separated class manifolds, but a lightweight student matching teacher relations only inherits an approximate version of that discriminative geometry. We therefore propose PDR, a feature-level objective that directly encourages the student's embedding to be discriminative (high inter-class separability and low intra-class variance) rather than fully relying on mimicry to supply this structure indirectly. Inspired by Fisher-style discriminative criteria \cite{tian2023grow, choi2025improving} but tailored for semantic segmentation, our formulation operates across all $K$ semantic classes with at least two labeled locations in the batch.

\subsubsection{Between-Class Separation}
For each semantic class $k \in \{1,\dots,K\}$, we compute a centroid vector $\mu_k$ by averaging student encoder features over spatial locations belonging to that class. We treat these centroids as class prototypes and enforce inter-class separability by penalizing prototype pairs that fall within a margin $m$ of each other:
\begin{equation}\mathcal{L}_{sep} = \frac{2}{K(K - 1)} \sum_{k < k^\prime} \max(0, m - \|\mu_{k} - \mu_{k^\prime}\|_2),
\end{equation}where $m$ is a fixed margin. This formulation encourages prototypes to be at least a minimum distance apart in the embedding space, preventing class collapse while maintaining separation across all semantic categories, unlike global mean-based scatter objectives that can be dominated by a subset of classes. This is particularly important for semantic segmentation, where balanced separation across classes directly affects boundary quality.

\subsubsection{Within-Class Compactness} 
To encourage compactness, we minimize the intra-class dispersion of features within each semantic category. Let $f_k$ denote the set of feature vectors assigned to class $k$, with centroid $\mu_k$. Their covariance matrix is
\begin{equation}\Sigma_k = \frac{1}{|f_k| - 1} \sum_{f \in f_k} (f - \mu_k)(f - \mu_k)^\top,
\end{equation}and we define the compactness loss as:
\begin{equation}\mathcal{L}_{comp} = \frac{1}{K} \sum_{k=1}^{K} \mathrm{tr}(\Sigma_k),
\end{equation}where $\mathrm{tr}(\cdot)$ is the matrix trace, so $\mathrm{tr}(\Sigma_k)$ equals the sum of per-feature variances within class $k$, encouraging compact feature clusters and stable prototype estimation.

\noindent\textbf{Combined PDR Objective.} The final objective integrates both terms to simultaneously encourage inter-class separation and intra-class compactness:
\begin{equation}\mathcal{L}_{PDR} = \lambda_{1} \mathcal{L}_{sep} + \lambda_{2} \mathcal{L}_{comp},
\end{equation}where $\lambda_1$ and $\lambda_2$ are hyperparameters that balance the two components. This formulation creates a complementary optimization dynamic that pushes class prototypes apart while tightening each class distribution. While conceptually related to Fisher-style criteria such as Linear Discriminant Analysis (LDA), our formulation differs in two important aspects: (i) it operates on batch-level statistics without requiring matrix inversion or eigen decomposition, making it computationally efficient and stable in high-dimensional settings; and (ii) it relies on pairwise prototype separation rather than the global mean-based between-class scatter ($S_B$), promoting more balanced class separation and avoiding the dominance by a subset of classes that can arise in $S_B$-based objectives. These properties make the objective particularly well-suited for semantic segmentation.

\subsection{Overall Objective}
The final training objective is:
\begin{equation}\mathcal{L}_{total} = \alpha\mathcal{L}_{task} + \beta\mathcal{L}_{MWAD} + \gamma\mathcal{L}_{PDR}\,,
\end{equation}\noindent where $\alpha$, $\beta$, $\gamma$ are hyperparameters balancing the three terms and $\mathcal{L}_{task}$ is the standard task loss (pixel-wise cross-entropy for segmentation).
In summary, the components described in this section form a coherent strategy for transformer-to-CNN distillation, with each design decision addressing a specific obstacle. (i) The channel projection and spatial resizing place heterogeneous teacher and student features on a common grid, a prerequisite for any meaningful comparison. (ii) The symmetric $1\times1$ QKV projections induce a shared relational view, since the CNN student does not natively produce attention but can be cast into the same relational basis as the transformer teacher through learned projections. (iii) Defining the distillation objective over relations (including attention maps, value embeddings, and aggregated contexts) rather than raw activations exploits the fact that relational structure transfers more readily across architectures than absolute feature values. (iv) Restricting attention to local windows brings the relational targets within the receptive-field reach of a convolutional student, while the multi-scale design recovers long-range structure at coarser stages where each window already spans a large image extent. (v) Stochastic shifts neutralize the arbitrariness of any particular window partition. (vi) PDR makes the class-wise geometry more compact and better separated than relation matching alone achieves under the student's limited capacity.

\section{Experiments and Results}

\subsection{Experimental Setup}
\paragraph{Datasets.}
We evaluate our approach on two distinct dense prediction domains: urban scene parsing and medical image segmentation.
For urban scene segmentation, we use the Cityscapes dataset \cite{cordts2016cityscapes}, which contains 5,000 finely annotated street-level images from 50 cities, with 2,975 training, 500 validation, and 1,525 test samples across 19 semantic classes. All images are dynamically cropped to $512 \times 1024$ for training and evaluated at full resolution using bilinear upsampling of the predicted masks.
For medical image segmentation, we use two polyp segmentation benchmarks: CVC-ClinicDB \cite{bernal2015wm} and CVC-ColonDB \cite{bernal2012towards}. CVC-ClinicDB contains 612 frames from 29 colonoscopy sequences, split into 80/10/10 for training, validation, and testing. CVC-ColonDB consists of 300 images from 15 sequences with higher intra-class variability, split into 240 training and 30 images each for validation and testing. All images are resized to $512 \times 512$.

\paragraph{Implementation Details.}
All experiments are implemented in PyTorch and conducted on a single NVIDIA A100 (80GB) GPU. To account for the differences between urban and medical domains, we adopt two distinct teacher-student configurations.
For Cityscapes, we use the frozen SAM2.1 Hiera-L backbone \cite{ravi2025sam} with a DeepLabV3+ segmentation head as the teacher. The student model is DeepLabV3+ \cite{chen2018encoder} with a MobileNetV3 backbone \cite{howard2019searching} (output stride 16). Training is performed for 40,000 iterations with a batch size of 16, using the AdamW optimizer \cite{loshchilov2017decoupled} with an initial learning rate of $1 \times 10^{-3}$ and weight decay of $1 \times 10^{-4}$, following a polynomial learning rate schedule.
For medical image segmentation, we use MedSAM2's \cite{ma2025medsam2} Hiera-T backbone with a ResUNet++ segmentation head as the frozen teacher. We evaluate two student architectures, ResUNet++ \cite{jha2019resunet++} and UNeXt \cite{valanarasu2022unext}, to demonstrate generality. Models are trained for 200 epochs with a batch size of 8 using the AdamW optimizer with an initial learning rate of $5 \times 10^{-4}$ and weight decay of $1 \times 10^{-4}$, with cosine annealing of the learning rate.
Across both domains, teachers remain fully frozen during distillation. For MWAD, SWARD aligns intermediate representations across $N = 4$ stages with window sizes $M_i = [32, 16, 8, 4]$ and QKV projection dimensions $d_i = C^i_s$. Stochastic shifts are sampled independently per stage from the integers in $[-M_i, M_i]$ each iteration.
The same loss weights are used in all experiments. In the overall objective, we set $\alpha = 1.0$, $\beta = 2.0$, and $\gamma = 1.0$ for the task, MWAD, and PDR terms, respectively. Within PDR, we set $\lambda_1 = \lambda_2 = 1.0$ with a separation margin of $m = 1.3$.

\subsection{Quantitative Results}
We compare SWARD against several state-of-the-art knowledge distillation methods for dense prediction under cross-architectural settings (i.e., CIRKD \cite{yang2022cross}, CKD \cite{liu2022cross}, MasKD \cite{huang2022masked}, OFA-KD \cite{hao2023one}, FreeKD \cite{zhang2024freekd}, and HMKD \cite{xu2025hmkd}).

\paragraph{Urban scene parsing.} Table \ref{tab:cityscapes_results} presents the quantitative results on the Cityscapes semantic segmentation benchmark. SWARD consistently achieves the best performance among all compared methods, reaching 69.97\% mIoU and 77.75\% mAcc and improving over the from-scratch student (65.05\% mIoU) by 4.92 points. The competing KD methods yield only modest gains in this heterogeneous setting; the strongest, HMKD, reaches 68.73\% mIoU, still trailing SWARD by 1.24 points.
\begin{table}[t]
\setlength\abovecaptionskip{2pt}
\centering
\small
\setlength{\tabcolsep}{3.5pt}
\begin{tabular}{l|c|c|cc|cc}
\hline
\multirow{2}{*}{Method} & \multirow{2}{*}{Params (M)} & \multirow{2}{*}{FLOPs (G)} & \multicolumn{2}{c|}{Performance (\%)} & \multirow{2}{*}{Lat. (ms) $\downarrow$} & \multirow{2}{*}{FPS $\uparrow$} \\
 & & & mIoU & mAcc & & \\ \hline
T: SAM2.1 Hiera-L & 224.45 & 406.01 & 74.61 & 84.01 & 253.97 & 3.94 \\ \hline
S: DeepLabV3+ - MBV3-S & \multirow{8}{*}{3.30} & \multirow{8}{*}{22.60} & 65.05 & 72.86 & \multirow{8}{*}{4.24} & \multirow{8}{*}{235.86} \\
\quad +CIRKD \cite{yang2022cross} & & & 66.12 & 73.61 & & \\
\quad +CKD \cite{liu2022cross} & & & 66.91 & 73.26 & & \\
\quad +MasKD \cite{huang2022masked} & & & 67.58 & 75.12 & & \\
\quad +OFA-KD \cite{hao2023one} & & & 66.84 & 72.98 & & \\
\quad +FreeKD \cite{zhang2024freekd} & & & 67.99 & 75.93 & & \\
\quad +HMKD \cite{xu2025hmkd} & & & 68.73 & 76.06 & & \\
\rowcolor{gray!20}
\quad +\textbf{SWARD (Ours)} & & & \textbf{69.97} & \textbf{77.75} & & \\
\hline
\end{tabular}
\caption{Comparison on the Cityscapes dataset using DeepLabV3+ with a SAM2.1 Hiera-L backbone as the teacher and DeepLabV3+ with a MobileNetV3 backbone as the student. All distillation modules are training-only and add zero inference overhead at test time. Peak allocated memory is $\sim1264.33$ MB for the teacher versus $\sim51.93$ MB for the student.}
\label{tab:cityscapes_results}
\end{table}
The superiority of SWARD comes from its design to bridge the architectural gap by transferring attention, value, and context relations within stochastically shifted, multi-scale local windows. The proposed MWAD module allows the student to capture the teacher's structured spatial dependencies within localized neighborhoods, while PDR improves the semantic organization of the student feature space (as will be shown later). Together, these components recover roughly half of the teacher-student mIoU gap (4.92 of 9.56 points) with only $\sim$1.5\% of the teacher's parameters and $\sim$5.6\% of its FLOPs, showcasing the effectiveness of SWARD for urban scene parsing.

\paragraph{Polyp segmentation.} Table \ref{tab:polyp_results} reports the results on the polyp segmentation benchmarks. SWARD achieves the strongest performance across both datasets and both student architectures. With a ResUNet++ student, SWARD reaches 94.45\% mDice on CVC-ClinicDB, essentially matching the teacher (94.60\%), and 88.49\% mDice on the more demanding CVC-ColonDB, a 20.44-point improvement over the from-scratch baseline (68.05\%). With the smaller UNeXt student, SWARD attains 93.23\% and 85.30\% mDice on the two benchmarks. Because CVC-ClinicDB and CVC-ColonDB are small, we verified the stability of SWARD across three random seeds.\footnote{Over 3 seeds, SWARD attains $94.02 \pm 0.44$ (ResUNet++) and $93.22 \pm 0.39$ (UNeXt) mDice on CVC-ClinicDB, and $88.24 \pm 0.31$ (ResUNet++) and $85.40 \pm 0.36$ (UNeXt) mDice on CVC-ColonDB. Tables report a single run.}
\begin{table}[t]
\setlength\abovecaptionskip{2pt}
\centering
\small
\setlength{\tabcolsep}{2.2pt}
\begin{tabular}{l|c|c|cc|cc|cc}
\hline
\multirow{2}{*}{Method} & \multirow{2}{*}{Params (M)} & \multirow{2}{*}{FLOPs (G)} & \multicolumn{2}{c|}{CVC-ClinicDB} & \multicolumn{2}{c|}{CVC-ColonDB} & \multirow{2}{*}{Lat. (ms) $\downarrow$} & \multirow{2}{*}{FPS $\uparrow$} \\
 & & & mDice & mIoU & mDice & mIoU & & \\ \hline
T: MedSAM2 & 38.96 & 55.38 & 94.60 & 90.29 & 93.03 & 87.36 & 23.34 & 42.85 \\ \hline
S: ResUNet++ & \multirow{8}{*}{4.06} & \multirow{8}{*}{63.42} & 89.53 & 85.27 & 68.05 & 57.96 & \multirow{8}{*}{9.29} & \multirow{8}{*}{107.68} \\
\quad +CIRKD \cite{yang2022cross} & & & 91.39 & 85.21 & 78.12 & 70.81 & & \\
\quad +CKD \cite{liu2022cross} & & & 92.08 & 86.93 & 79.57 & 71.10 & & \\
\quad +MasKD \cite{huang2022masked} & & & 90.62 & 84.32 & 77.52 & 69.83 & & \\
\quad +OFA-KD \cite{hao2023one} & & & 91.98 & 86.02 & 78.18 & 70.86 & & \\
\quad +FreeKD \cite{zhang2024freekd} & & & 91.61 & 85.77 & 85.92 & 78.57 & & \\
\quad +HMKD \cite{xu2025hmkd} & & & 93.14 & 89.53 & 86.92 & 79.09 & & \\
\rowcolor{gray!20}
\quad +\textbf{SWARD (Ours)} & & & \textbf{94.45} & \textbf{90.03} & \textbf{88.49} & \textbf{80.90} & & \\ \hline
S: UNeXt & \multirow{8}{*}{1.47} & \multirow{8}{*}{4.63} & 90.02 & 84.21 & 80.65 & 72.70 & \multirow{8}{*}{3.52} & \multirow{8}{*}{284.09} \\
\quad +CIRKD \cite{yang2022cross} & & & 91.07 & 84.85 & 83.08 & 74.16 & & \\
\quad +CKD \cite{liu2022cross} & & & 91.91 & 86.26 & 83.41 & 74.87 & & \\
\quad +MasKD \cite{huang2022masked} & & & 91.72 & 86.01 & 82.83 & 73.09 & & \\
\quad +OFA-KD \cite{hao2023one} & & & 92.08 & 86.39 & 83.89 & 75.07 & & \\
\quad +FreeKD \cite{zhang2024freekd} & & & 91.51 & 85.18 & 83.17 & 74.58 & & \\
\quad +HMKD \cite{xu2025hmkd} & & & 92.96 & 87.19 & 84.72 & 76.99 & & \\
\rowcolor{gray!20}
\quad +\textbf{SWARD (Ours)} & & & \textbf{93.23} & \textbf{87.78} & \textbf{85.30} & \textbf{77.42} & & \\ \hline
\end{tabular}
\vspace{-12pt}
\caption{Comparison on polyp segmentation datasets using student-specific segmentation heads with a MedSAM2 Hiera-T backbone as the teacher and ResUNet++ and UNeXt as the students. All distillation modules are training-only and add zero inference overhead. Peak allocated memory is $\sim612.21$ MB for the teacher versus $\sim355.71$ MB for ResUNet++ and $\sim42.18$ MB for UNeXt students.}
\label{tab:polyp_results}
\end{table}
These consistent gains reflect the effectiveness of SWARD in transferring both relational and discriminative knowledge across student architectures (ResUNet++, UNeXt) and dataset difficulties (CVC-ClinicDB, CVC-ColonDB). Notably, ResUNet++ reaches teacher-level mDice on CVC-ClinicDB with only $\sim$10\% of the teacher's parameters, while the lighter UNeXt achieves competitive accuracy with $\sim$3.8\% of the teacher's parameters and $\sim$8.4\% of its FLOPs.

\subsection{Qualitative Results}
\subsubsection{Segmentation Predictions}
\begin{figure}[t]
    \centering
    \makebox[0.16\textwidth][c]{\small Input Image}\hfill
    \makebox[0.16\textwidth][c]{\small Ground Truth}\hfill
    \makebox[0.16\textwidth][c]{\small Teacher}\hfill
    \makebox[0.16\textwidth][c]{\small Baseline}\hfill
    \makebox[0.16\textwidth][c]{\small FreeKD}\hfill
    \makebox[0.16\textwidth][c]{\small SWARD (Ours)}

    \makebox[0.16\textwidth][c]{\subfigure{\includegraphics[width=0.166\textwidth]{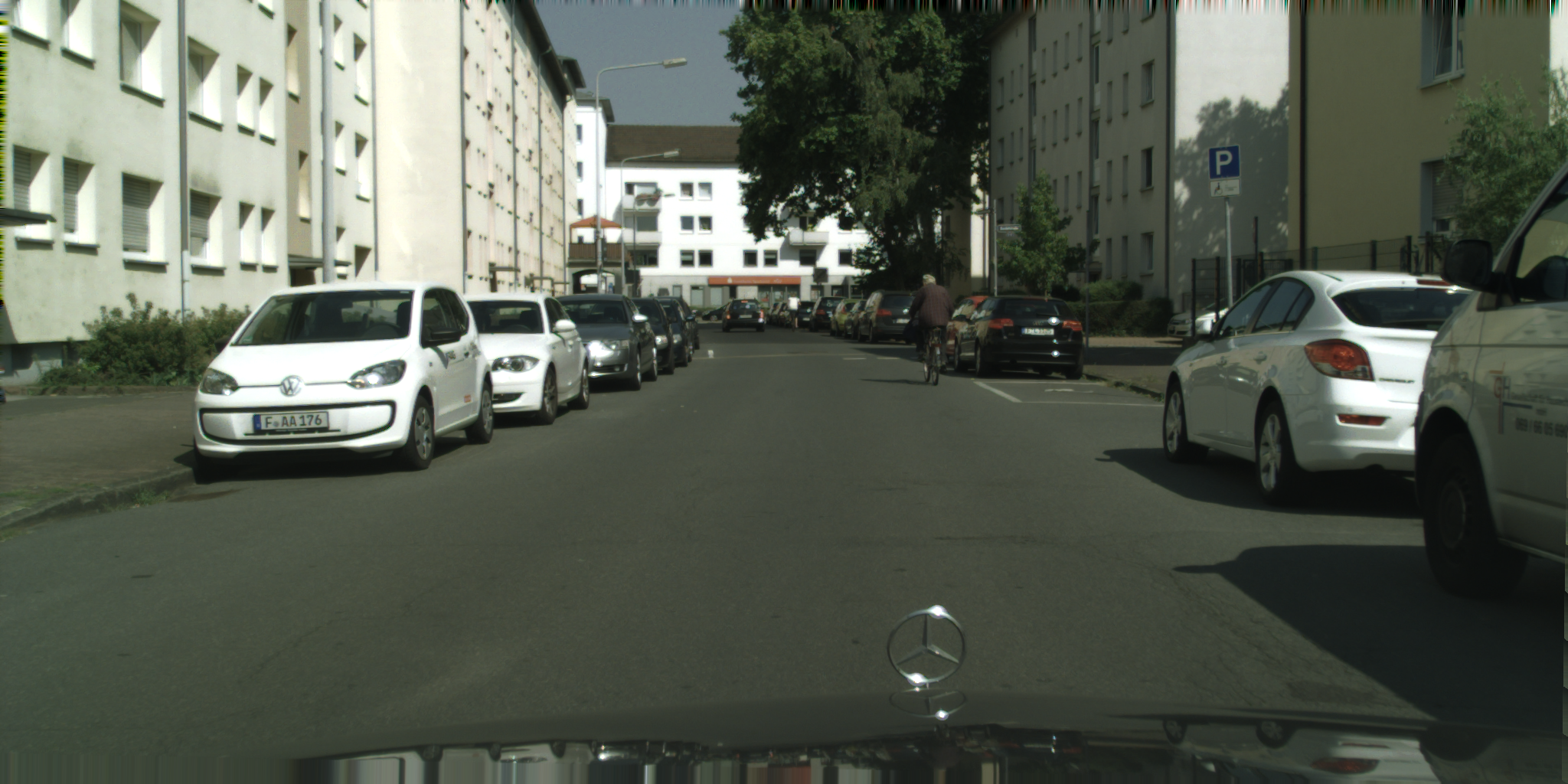}}}\hfill
    \makebox[0.16\textwidth][c]{\subfigure{\includegraphics[width=0.166\textwidth]{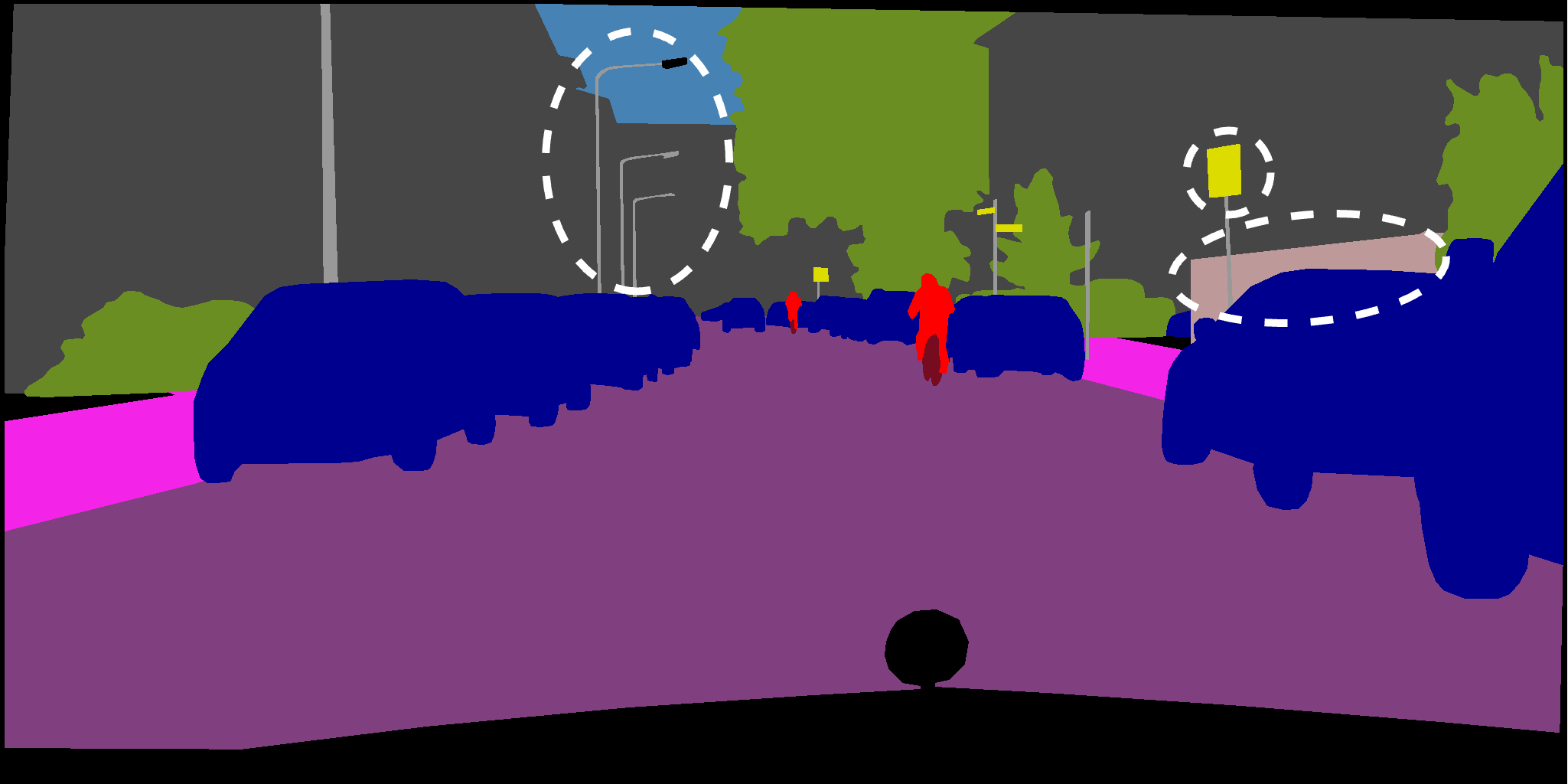}}}\hfill
    \makebox[0.16\textwidth][c]{\subfigure{\includegraphics[width=0.166\textwidth]{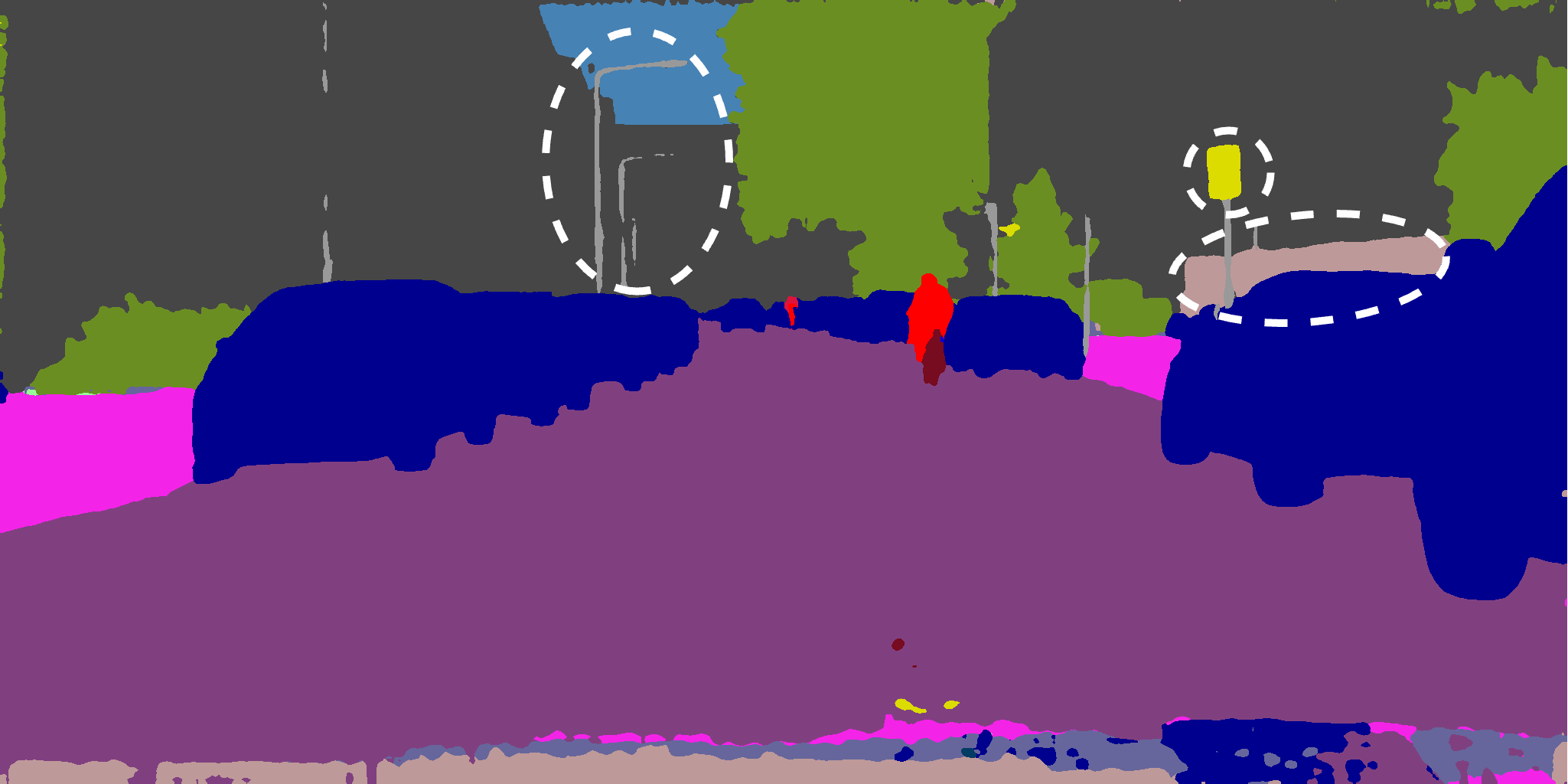}}}\hfill
    \makebox[0.16\textwidth][c]{\subfigure{\includegraphics[width=0.166\textwidth]{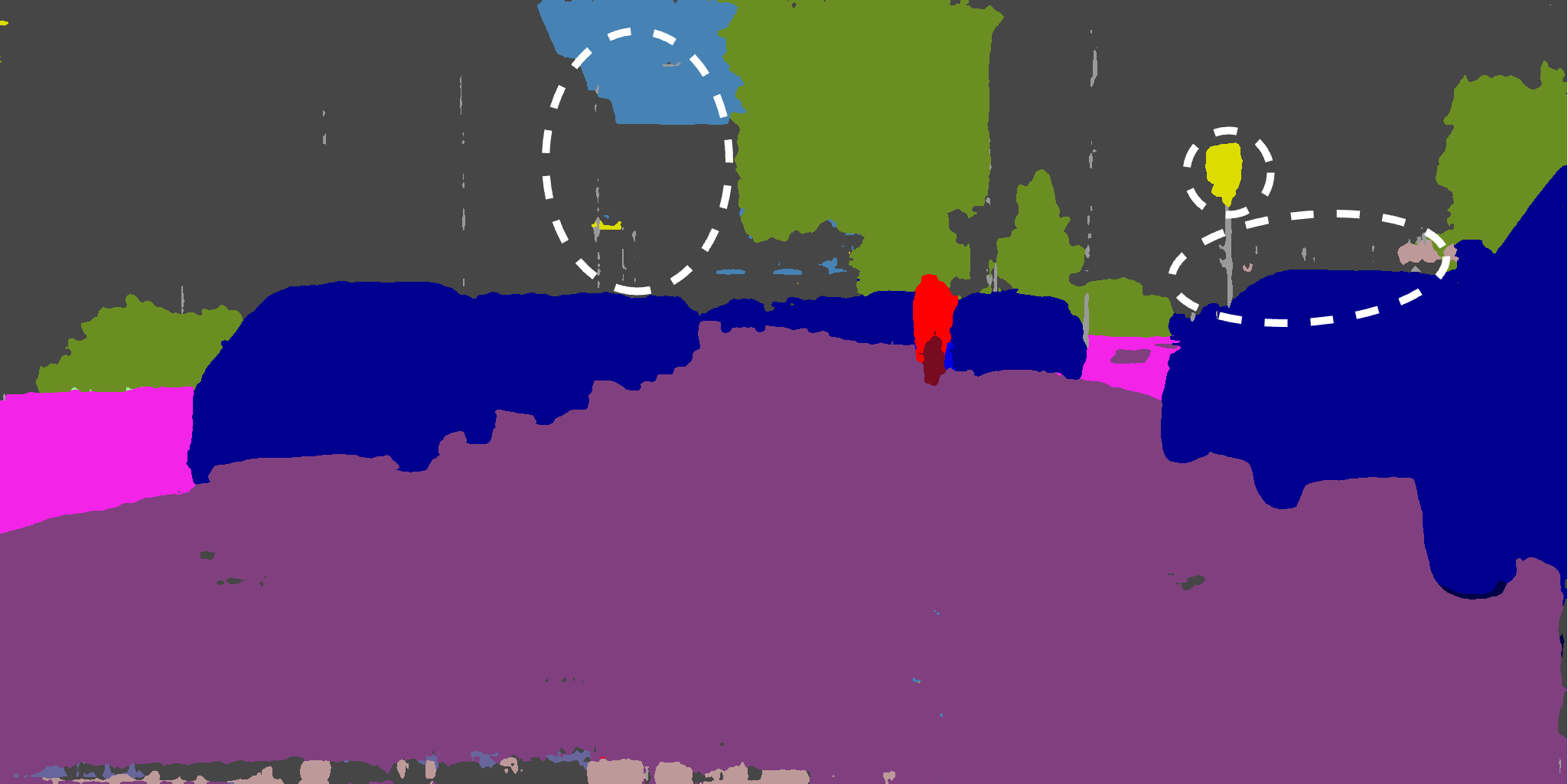}}}\hfill
    \makebox[0.16\textwidth][c]{\subfigure{\includegraphics[width=0.166\textwidth]{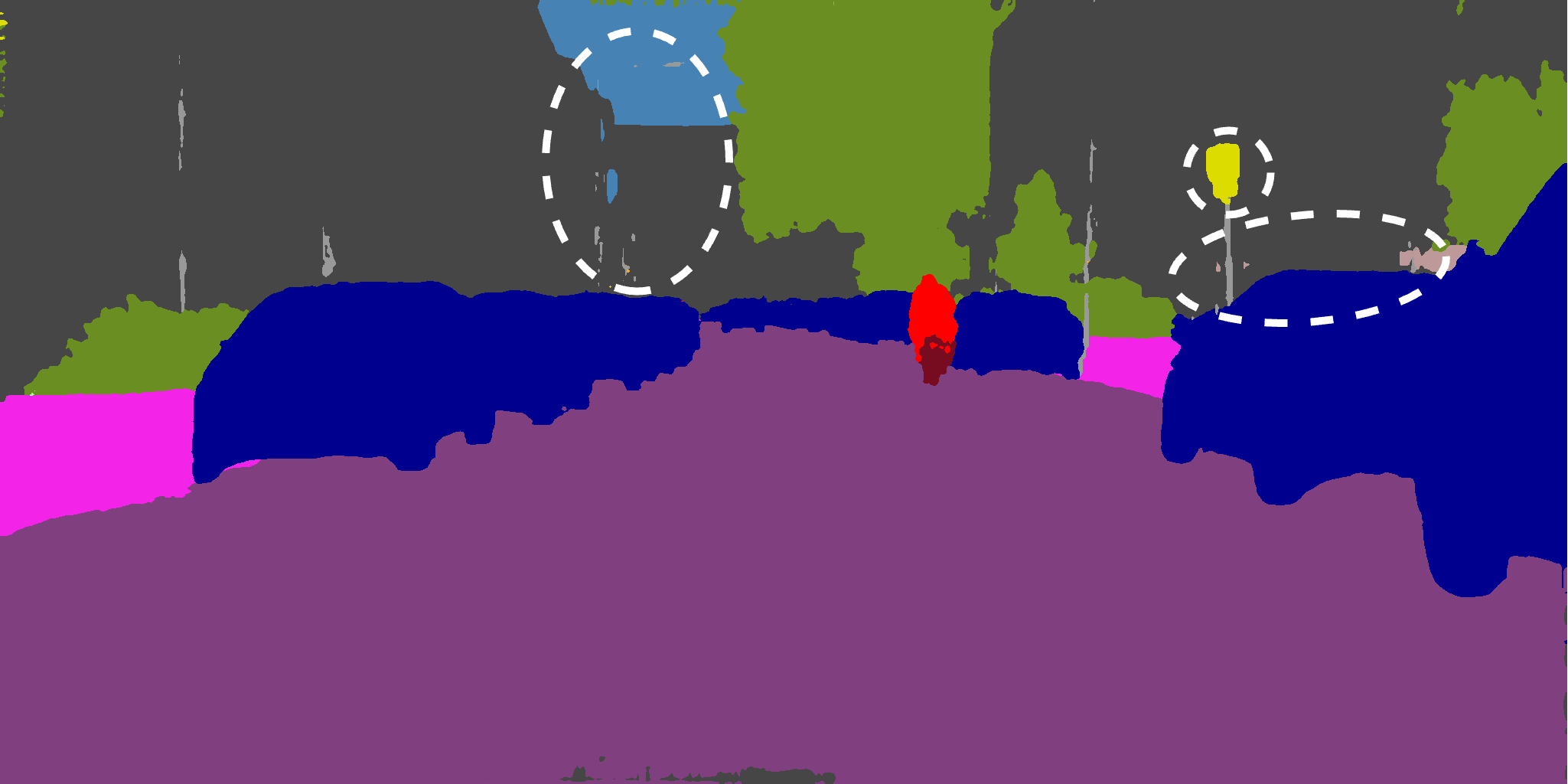}}}\hfill
    \makebox[0.16\textwidth][c]{\subfigure{\includegraphics[width=0.166\textwidth]{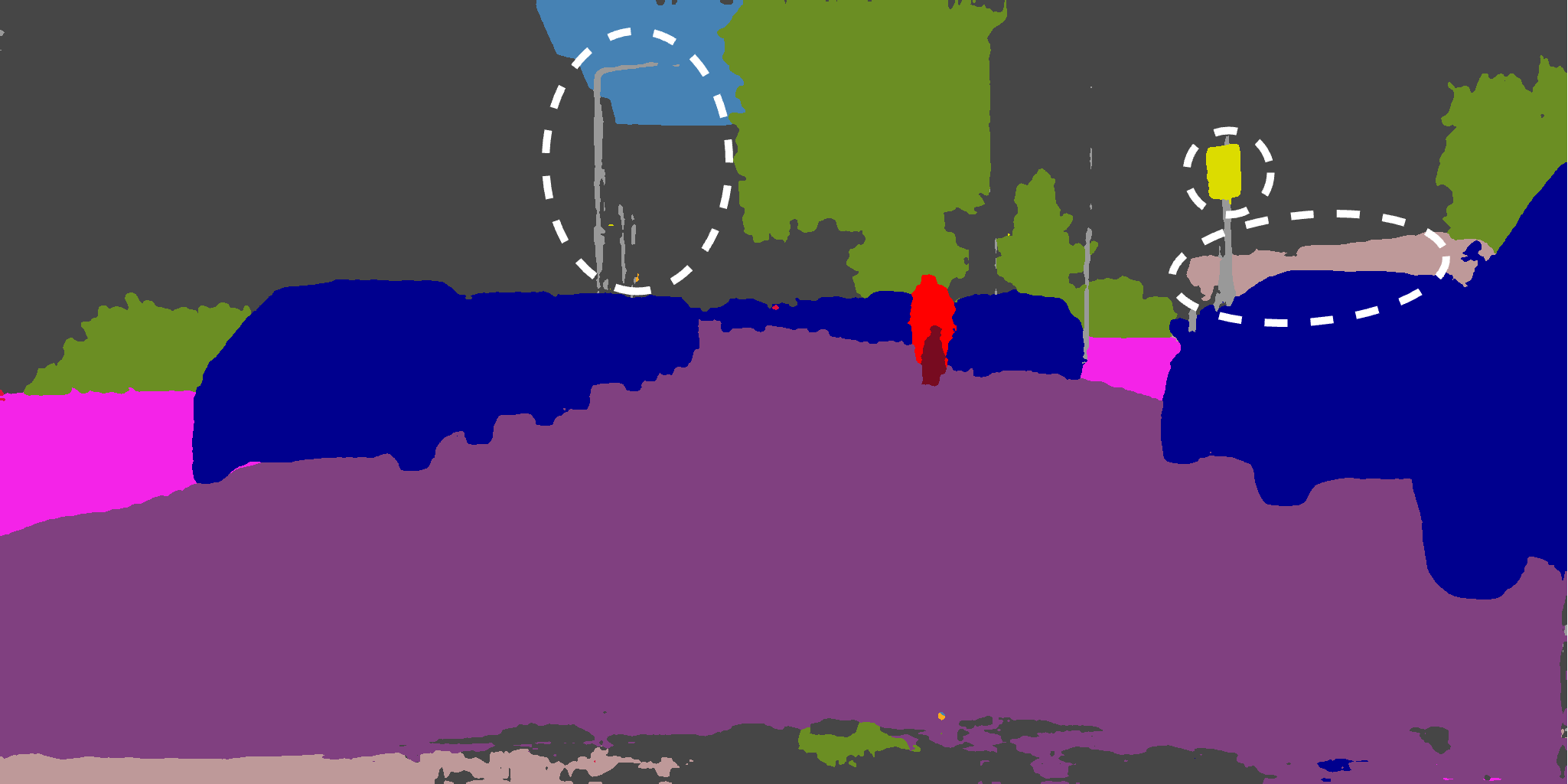}}}
    
    \par\vspace{-10pt}

    \makebox[0.16\textwidth][c]{\subfigure{\includegraphics[width=0.13\textwidth]{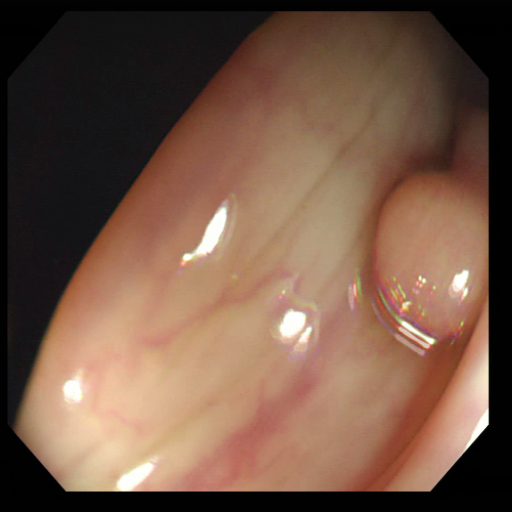}}}\hfill
    \makebox[0.16\textwidth][c]{\subfigure{\includegraphics[width=0.13\textwidth]{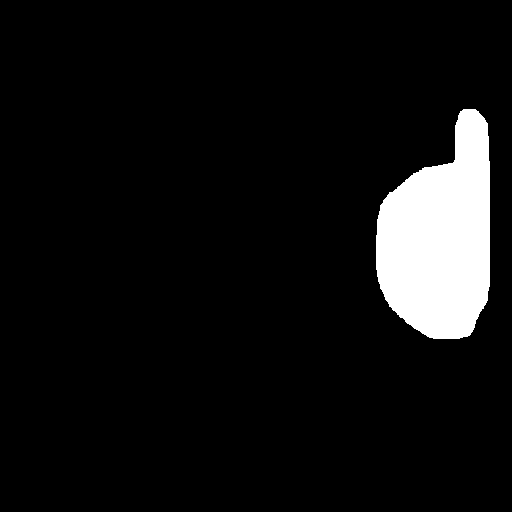}}}\hfill
    \makebox[0.16\textwidth][c]{\subfigure{\includegraphics[width=0.13\textwidth]{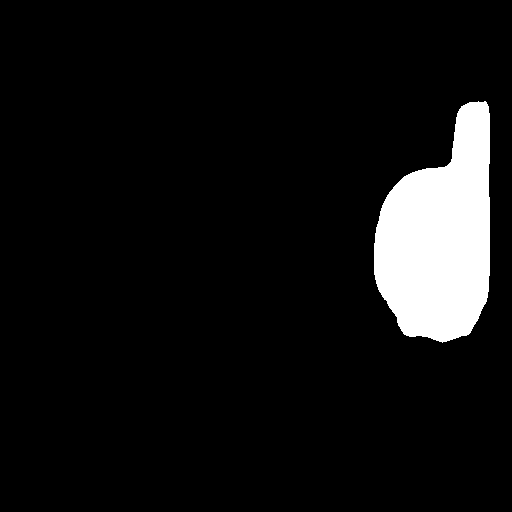}}}\hfill
    \makebox[0.16\textwidth][c]{\subfigure{\includegraphics[width=0.13\textwidth]{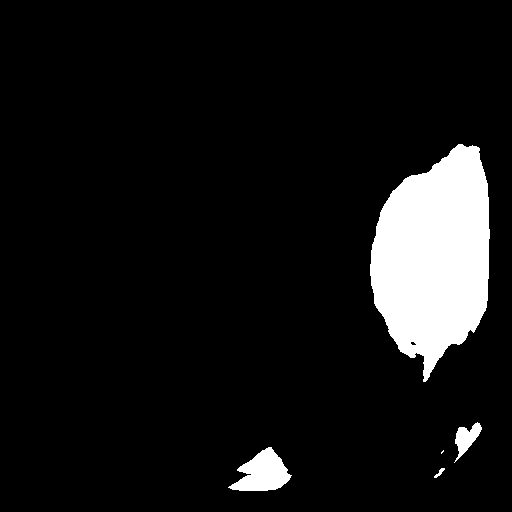}}}\hfill
    \makebox[0.16\textwidth][c]{\subfigure{\includegraphics[width=0.13\textwidth]{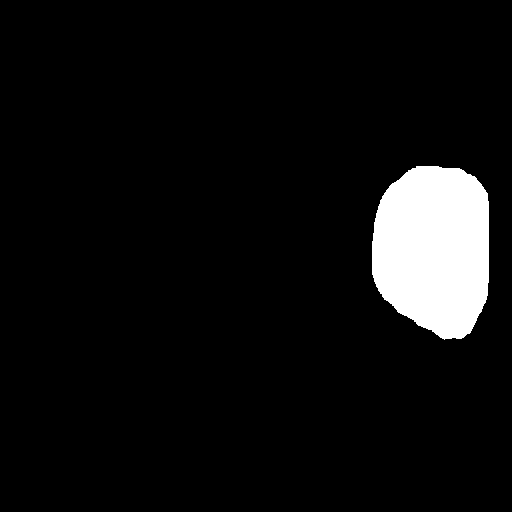}}}\hfill
    \makebox[0.16\textwidth][c]{\subfigure{\includegraphics[width=0.13\textwidth]{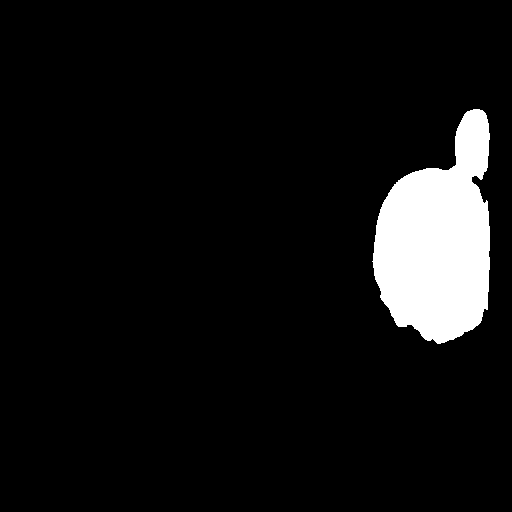}}}
    
    \par\vspace{-10pt}

    \makebox[0.16\textwidth][c]{\subfigure{\includegraphics[width=0.13\textwidth]{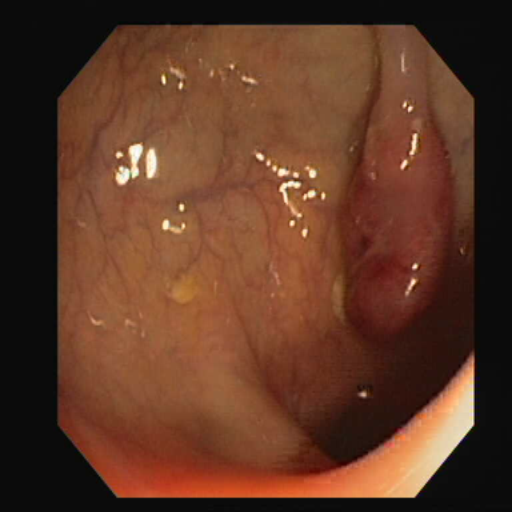}}}\hfill
    \makebox[0.16\textwidth][c]{\subfigure{\includegraphics[width=0.13\textwidth]{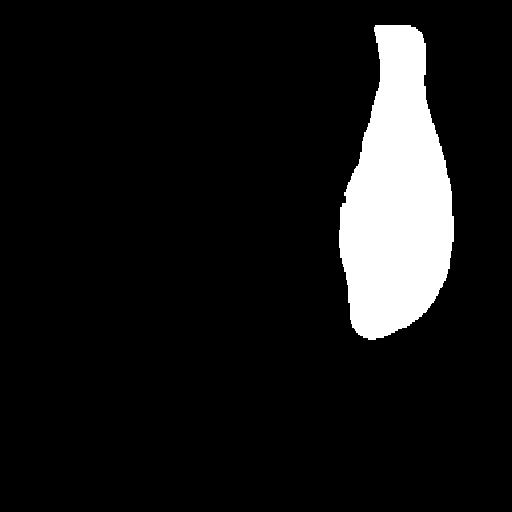}}}\hfill
    \makebox[0.16\textwidth][c]{\subfigure{\includegraphics[width=0.13\textwidth]{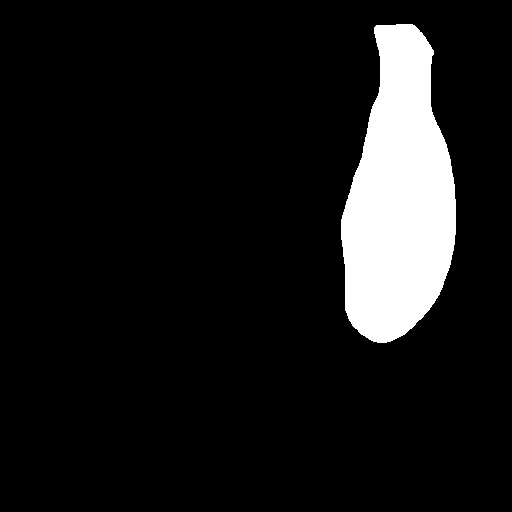}}}\hfill
    \makebox[0.16\textwidth][c]{\subfigure{\includegraphics[width=0.13\textwidth]{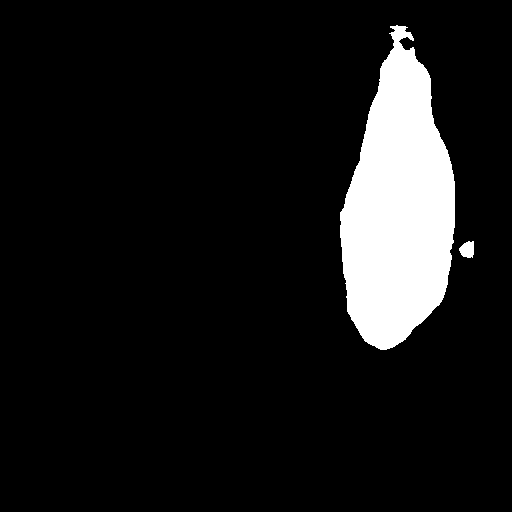}}}\hfill
    \makebox[0.16\textwidth][c]{\subfigure{\includegraphics[width=0.13\textwidth]{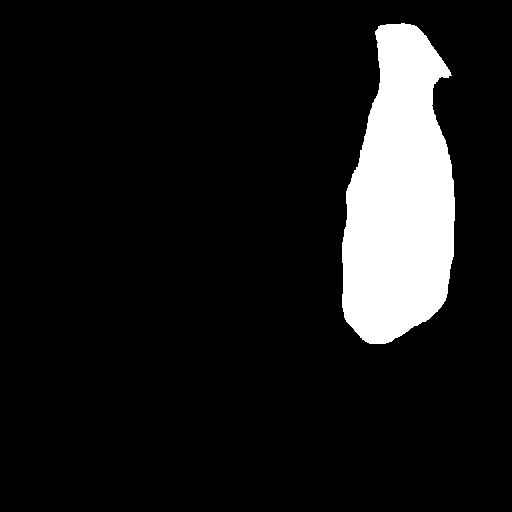}}}\hfill
    \makebox[0.16\textwidth][c]{\subfigure{\includegraphics[width=0.13\textwidth]{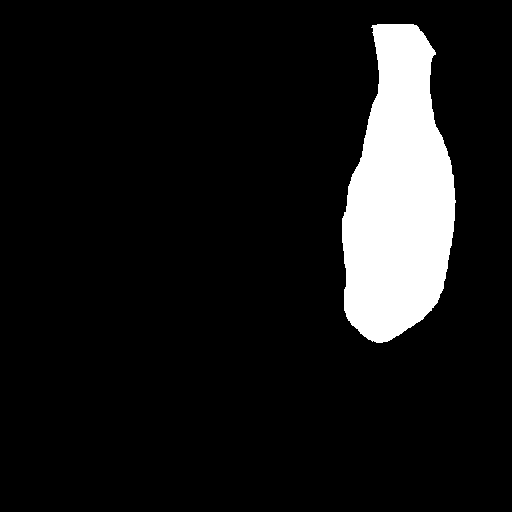}}}

    \par\vspace{-6pt}

    \caption{Comparison of segmentation predictions on Cityscapes and polyp datasets. Rows from top to bottom: Cityscapes, CVC-ColonDB, and CVC-ClinicDB; columns from left to right: input image, ground-truth mask, teacher prediction, baseline student, +FreeKD, and +SWARD (ours). Dashed ellipses in the first row highlight regions where the methods differ.}
    \label{fig:pred_masks_comparison}
    \vspace{-10pt}
\end{figure}

As shown in Figure \ref{fig:pred_masks_comparison}, the baseline student frequently produces fragmented predictions and inconsistent object boundaries, particularly around thin structures and small semantic regions.
On the Cityscapes example, the three dashed ellipses mark a light pole, a traffic sign, and a fence. In each, the baseline produces noisy or missing predictions, FreeKD is slightly better but still fails to recover the correct shape, and SWARD reproduces the teacher's predictions closely.
On the polyp examples, the baseline produces jagged or fragmented boundaries (with additional false-positive blobs below the smaller lesion), FreeKD smooths these artifacts but loses fine shape details, omitting the smaller lesion's elongated apex entirely, and SWARD preserves the lesions' shapes and matches the teacher closely on both samples.

\subsubsection{Feature Space Visualization} 
\begin{figure}[t]
\centering
\subfigure[Baseline Student]{
    \includegraphics[width=0.22\linewidth]{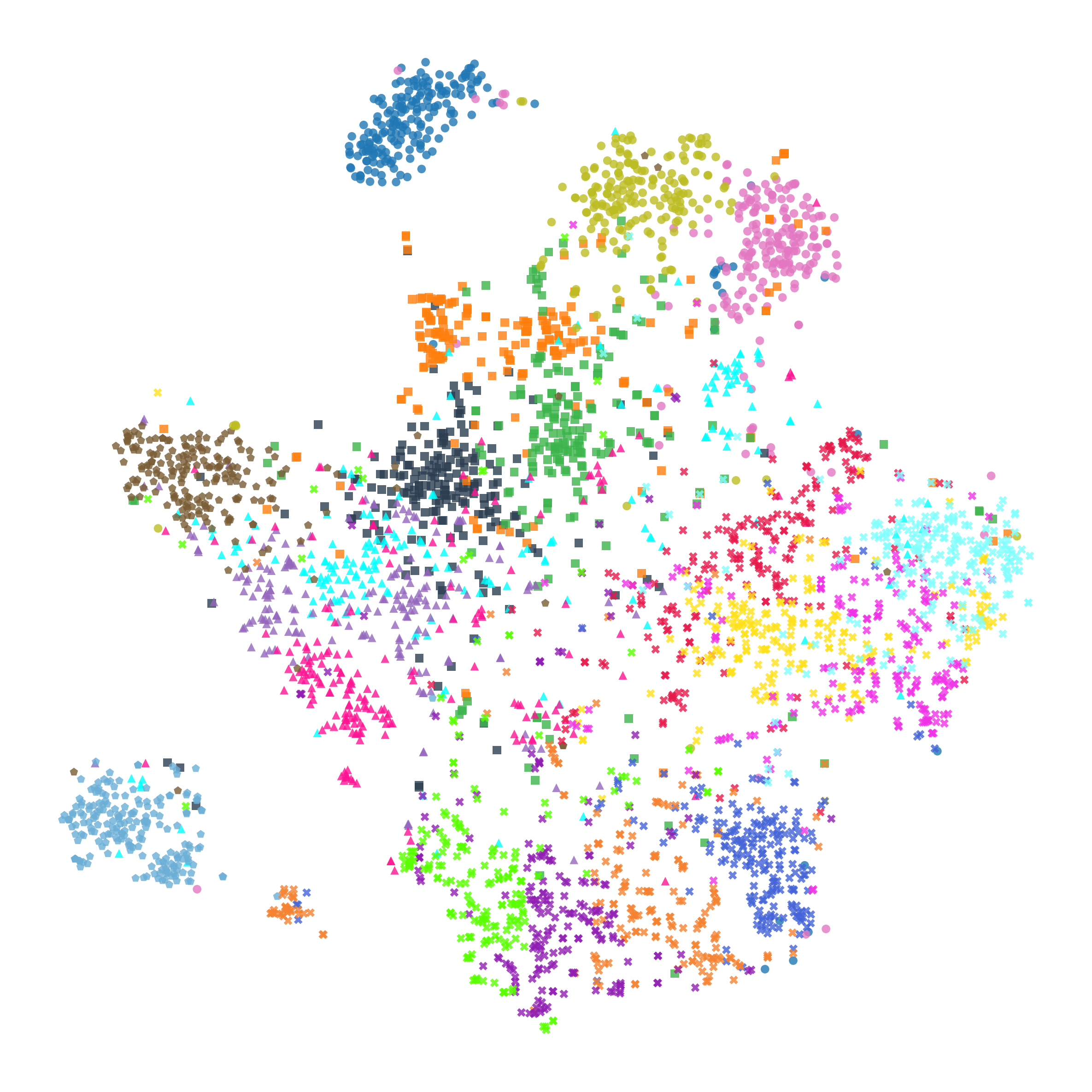}
}
\hfill
\subfigure[Teacher Network]{
    \includegraphics[width=0.22\linewidth]{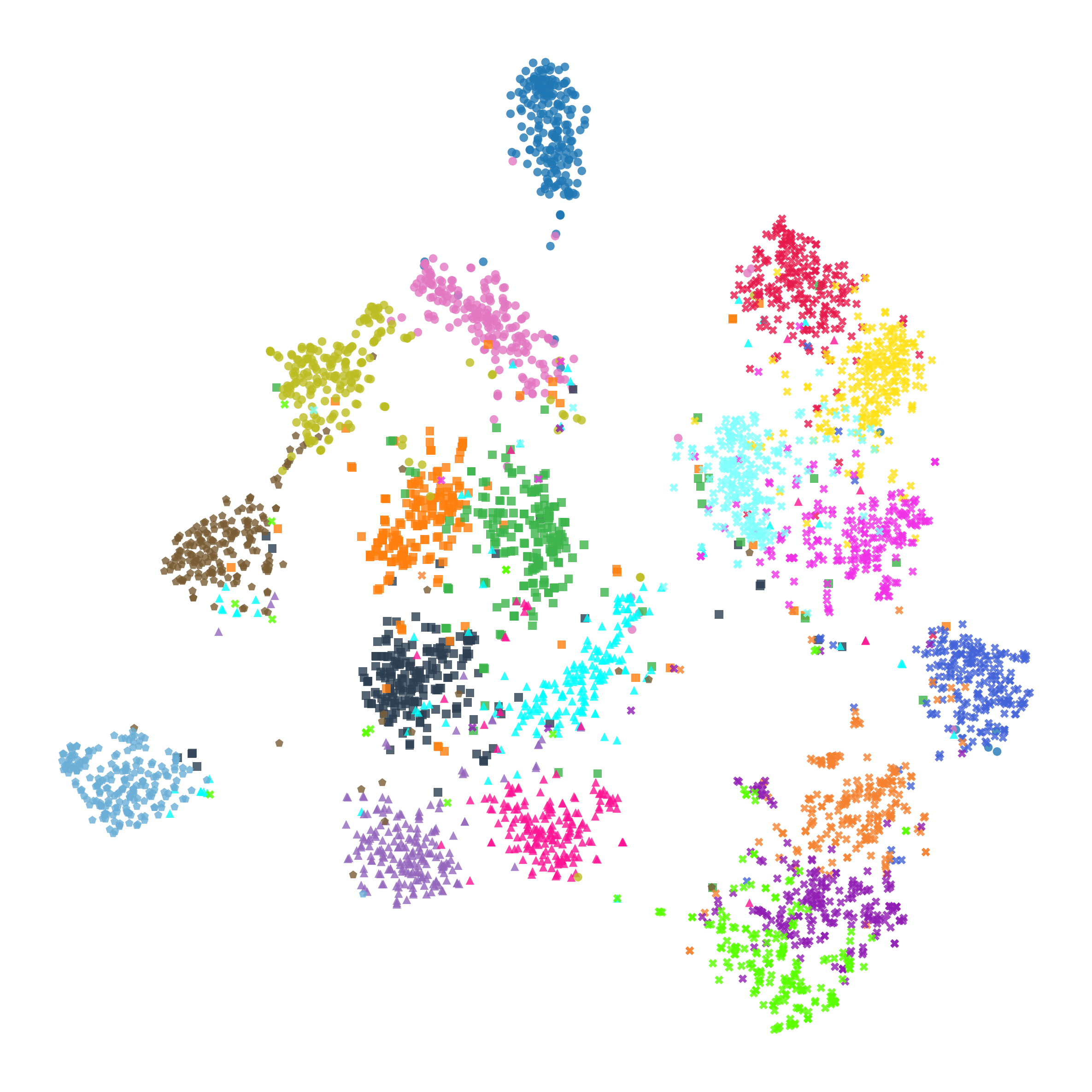}
}
\hfill
\subfigure[Student + MWAD]{
    \includegraphics[width=0.22\linewidth]{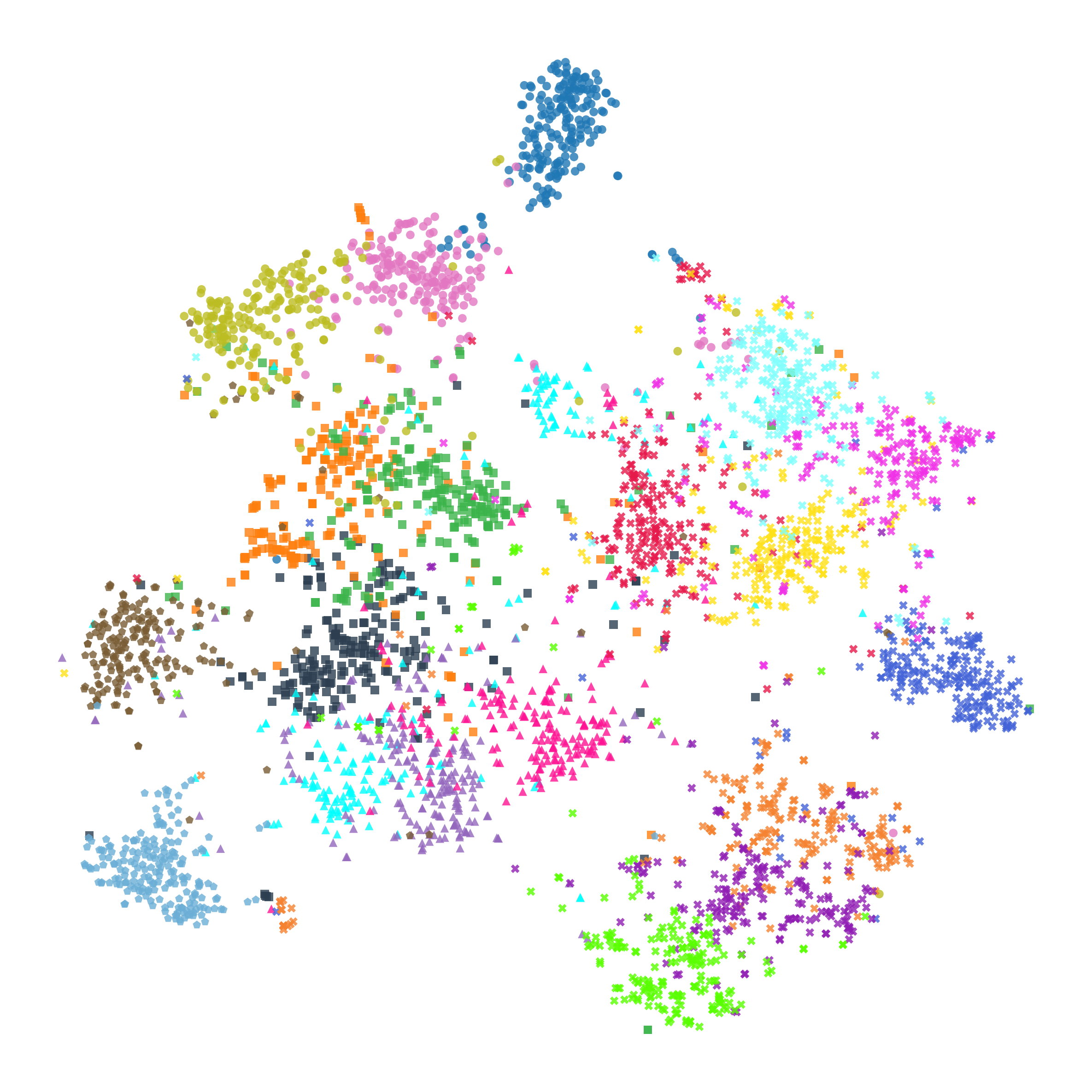}
}
\hfill
\subfigure[Student (SWARD)]{
    \includegraphics[width=0.22\linewidth]{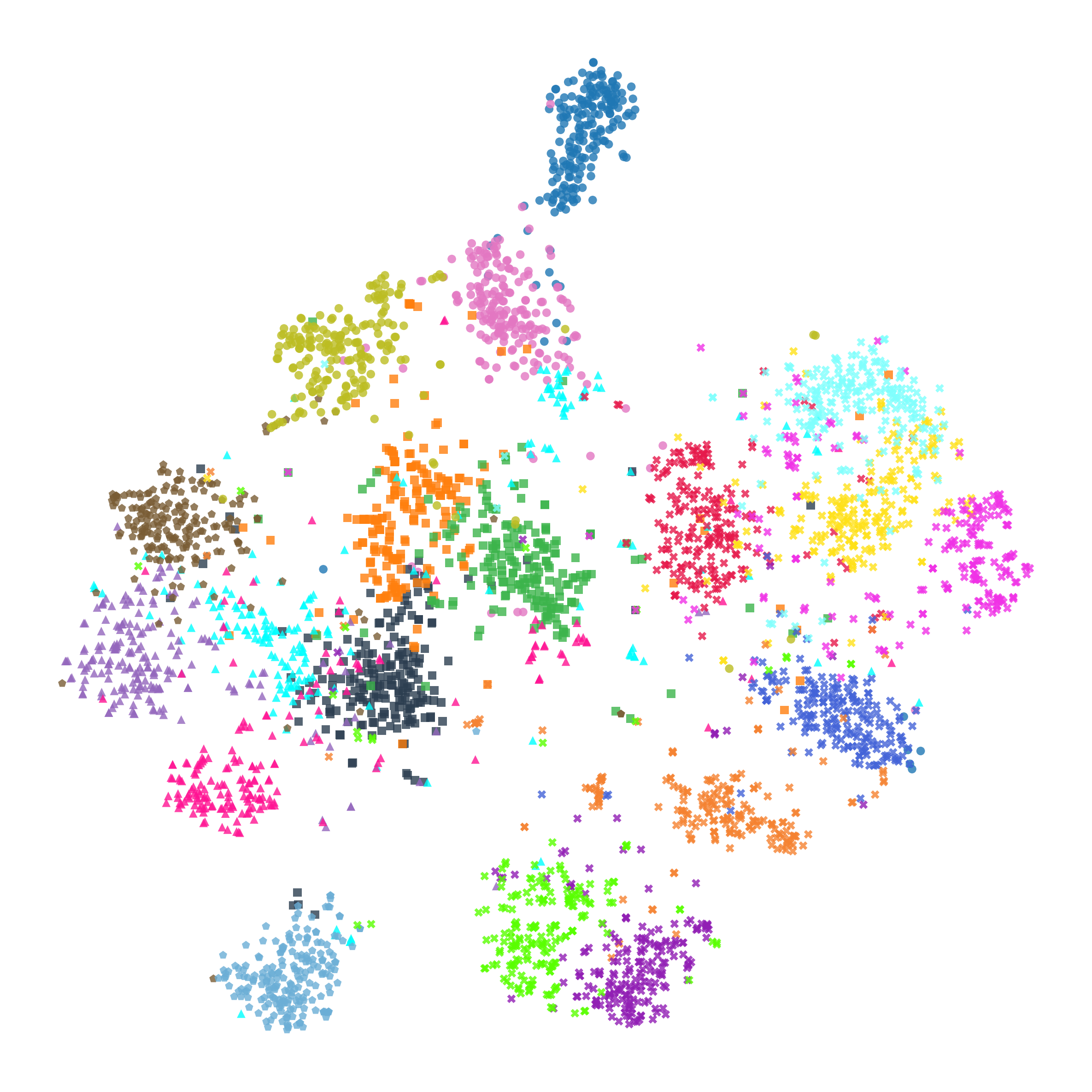}
}
\par\vspace{3pt}
\noindent\includegraphics[width=1.0\linewidth, trim={0 22 0 21}, clip]{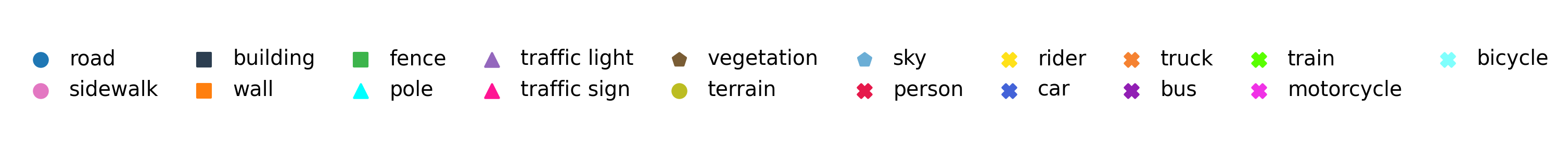}\par
\caption{Feature space visualization via $t$-SNE of penultimate-layer features on the Cityscapes validation set (colors denote classes; legend included). The baseline student's classes are heavily interleaved; MWAD transfers the teacher's cluster arrangement to the student, and adding PDR further separates and compacts the class clusters, reaching separability comparable to the teacher's.}
\label{fig:t_sne_plots}
\end{figure}

Figure \ref{fig:t_sne_plots} shows the $t$-SNE \cite{van2008visualizing} plots of the penultimate latent space features for the baseline student, teacher, student with MWAD, and student with MWAD + PDR (i.e., SWARD) models. In the student baseline's feature space, structurally and semantically similar categories are heavily interleaved. Classes like person (\mkperson), rider (\mkrider), motorcycle (\mkmotorcycle), and bicycle (\mkbicycle) form a highly diffused cluster, representing extreme feature overlap that leads to both inter-class misclassifications and incorrect segmentations.
With MWAD, the student's feature space becomes closer to the teacher's. Background classes like terrain (\mkterrain), sidewalk (\mksidewalk), and road (\mkroad), as well as the overall cluster arrangement, mirror the teacher's manifold, reflecting that MWAD effectively transfers the teacher's relational structure to the student.

Building upon these favorable alignment patterns, PDR introduces targeted refinement to the cluster geometry. While largely maintaining the structure inherited from the teacher, PDR explicitly enforces inter-class separation and intra-class compactness. The clusters in the MWAD+PDR plot become visibly better separated compared to MWAD alone (e.g., several clusters on the right side). Notably, with PDR the student attains class separation comparable to the teacher's despite its smaller capacity. For several visually or semantically similar categories, e.g., truck (\mktruck) vs. bus (\mkbus), bicycle (\mkbicycle) vs. motorcycle (\mkmotorcycle), the MWAD+PDR student not only exhibits cleaner inter-class separation than other students, but also achieves even cleaner separation than the teacher itself. This advantage over the teacher is also evident for vegetation (\mkvegetation) versus terrain (\mkterrain), where the teacher's own clusters remain partially intermingled while the MWAD+PDR student forms cleanly separated regions.

\subsection{Ablation Studies}
To better understand the contribution of each component in SWARD, we conduct a series of ablation studies on the CVC-ColonDB benchmark. Compared to CVC-ClinicDB, ColonDB contains smaller polyps, lower contrast boundaries, and more diverse appearances, making it a more demanding testbed for evaluating the effectiveness of each design choice.

\begin{table}[h]
\setlength\abovecaptionskip{2pt}
\centering
\begin{tabular}{l|cc}
\hline
\multirow{2}{*}{Method} & \multicolumn{2}{c}{Performance (\%)} \\
 & mDice & mIoU \\ \hline
T: MedSAM2 & 93.03 & 87.36 \\ \hline
S: ResUNet++ & 68.05 & 57.96 \\ \hline
+ MWAD (Attention only) & 73.19 & 63.77 \\
+ MWAD (Attn + Value) & 80.17 & 72.04 \\
+ MWAD (Attn + Value + Context) & 87.02 & 79.58 \\ \hline
\rowcolor{gray!20}
+ MWAD + PDR & \textbf{88.49} & \textbf{80.90} \\
\hline
\end{tabular}
\caption{Ablation study of SWARD components on the CVC-ColonDB dataset with MedSAM2 as the teacher and ResUNet++ as the student. We progressively add components of MWAD and evaluate the final combination with PDR.}
\label{tab:colondb_ablation}
\end{table}

We first evaluate the contribution of the MWAD module by progressively adding its components: attention alignment, value alignment, and context alignment. As shown in Table \ref{tab:colondb_ablation}, attention alignment alone lifts the ResUNet++ student from $68.05\%$ to $73.19\%$ mDice, indicating that transferring the local relational structure from the teacher is highly effective. Adding value alignment further boosts the mDice to $80.17\%$ by enforcing semantic consistency between teacher and student features, and adding context alignment brings another $+6.85$ points by aligning aggregated representations. Together, the three terms recover about $76\%$ of the teacher-student mDice gap before PDR is applied.
Finally, adding the PDR loss to the MWAD objective lifts mDice to $88.49\%$, the best overall result. These gains show that MWAD and PDR provide complementary benefits. MWAD improves the student by transferring relational reasoning of the teacher through attention, value, and context alignment, while PDR shapes the student's feature space into well-separated, compact class clusters, adding the discriminative structure that relation matching alone cannot recover under the student's reduced capacity. 

\begin{table}[h]
\setlength\abovecaptionskip{2pt}
\centering
\begin{tabular}{l|cc}
\hline
\multirow{2}{*}{Window Strategy} & \multicolumn{2}{c}{Performance (\%)} \\
 & mDice & mIoU \\ \hline
No Shifting & 86.17 & 78.83 \\ \hline
Deterministic Shifting & 87.91 & 79.99 \\ \hline
\rowcolor{gray!20}
Stochastic Shifting & \textbf{88.49} & \textbf{80.90} \\
\hline
\end{tabular}
\caption{Ablation study of different window strategies on CVC-ColonDB with MedSAM2 as the teacher and ResUNet++ as the student.}
\label{tab:window_strategy_ablation}
\end{table}

To further study the effect of the window partitioning strategy, we compare three variants within MWAD: no shifting, deterministic shifting, and stochastic shifting, as shown in Table \ref{tab:window_strategy_ablation}. The no-shifting baseline reaches $86.17\%$ mDice. Its rigid spatial partitioning limits interaction across adjacent regions and can lead to fragmented feature learning at object boundaries. Introducing deterministic shifting \cite{liu2021swin} mitigates this issue by relocating window boundaries so that previously separated regions share a window, enabling cross-window information exchange and improving mDice by $+1.74$ points to $87.91\%$. However, the shift offset is fixed across both iterations and stages, constraining the relational contexts seen during training.
In contrast, our stochastic shifting reaches $88.49\%$ mDice, the best result overall. By resampling the window offset for each scale at every iteration, the student is exposed to a continually varying set of spatial decompositions of the same features. This both regularizes against overfitting to a single window alignment and enriches the relational distillation signal itself.

\section{Conclusion}
In this paper, we introduced SWARD, a knowledge distillation framework that transfers relational reasoning from a strong transformer-based teacher to a lightweight CNN-based student for semantic segmentation. SWARD decomposes knowledge transfer into attention, value, and context alignment within a windowed attention formulation, enabling the student to learn both the feature representations and the underlying spatial dependencies that drive the teacher's predictions. The window partitions are stochastically shifted at every training iteration and scale, removing window-boundary bias and exposing the student to diverse spatial decompositions of the same features. As a complementary objective, we proposed Prototype Discriminative Regularization (PDR), a loss that shapes the student's feature distribution into well-separated, compact class clusters via prototype-based inter-class separation and intra-class compactness terms. PDR adds discriminative structure that relation matching alone leaves incomplete under the student's reduced capacity. Experiments on Cityscapes and the polyp segmentation benchmarks (CVC-ClinicDB and CVC-ColonDB) show that SWARD consistently outperforms state-of-the-art distillation baselines, surpassing the strongest by $+1.24$ mIoU on Cityscapes while using only $\sim$1.5\% of the teacher's parameters and $\sim$5.6\% of its FLOPs, and ablation studies confirm the contribution of each component. Overall, SWARD bridges CNN-student and attention-teacher architectures by combining structured relational distillation with prototype-based feature shaping.

\section*{Acknowledgment}
This work was supported by the National Science Foundation (NSF) under Award No. 2412285.

\bibliography{egbib}
\end{document}